\documentclass[11pt]{amsart}
\usepackage{graphicx} 
\usepackage{amsmath,amsfonts,amsthm,amssymb,amscd}
\usepackage{lipsum}
\usepackage{xcolor}
\usepackage{amsfonts}
\usepackage{graphicx}
\usepackage{epstopdf}
\usepackage{algorithm}
\usepackage{algorithmic}
\usepackage{upgreek}
\usepackage{overpic}
\usepackage{enumerate}
\usepackage{enumitem}
\usepackage{bm}
\usepackage{booktabs}
\usepackage[foot]{amsaddr}
\usepackage[hidelinks]{hyperref}

\hypersetup{
  colorlinks=true,
  linkcolor=black,     
  citecolor=blue,      
  urlcolor=black,      
  filecolor=black      
}
\usepackage[letterpaper,top=3.5cm,bottom=3.5cm,left=3.5cm,right=3.5cm,marginparwidth=1.75cm]{geometry}

\newcommand{\cI}{\mathcal{I}}

\newcommand{\bE}{\mathbb{E}}

\newcommand{\bR}{\mathbb{R}}

\newcommand{\rd}{\mathrm{d}}

\newcommand{\sfN}{\mathsf{N}}

\newtheorem{theorem}{Theorem}[section]

\newtheorem{proposition}[theorem]{Proposition}
\newtheorem{definition}[theorem]{Definition}
\newtheorem{example}[theorem]{Example}

\theoremstyle{remark}
\newtheorem{remark}[theorem]{Remark}
\numberwithin{equation}{section}

\newcommand{\xqed}[1]{%
    \leavevmode\unskip\penalty9999 \hbox{}\nobreak\hfill
    \quad\hbox{\ensuremath{#1}}}
\newcommand{\Endofdef}{\xqed{\lozenge}}

\makeatletter
\renewcommand\paragraph{\@startsection{paragraph}{4}{\z@}%
  {-2.5ex\@plus -1ex \@minus -.25ex}%
  {-1em}%
  {\normalfont\normalsize\bfseries}}
\makeatother

\title[Scale-Adaptive Generative Flows for Multiscale Scientific Data]{Scale-Adaptive Generative Flows for Multiscale Scientific Data}

\author{Yifan Chen\textsuperscript{1}}
\address{\textsuperscript{1}Department of Mathematics, University of California, Los Angeles, CA, USA}
\email{yifanchen@math.ucla.edu}
\author{Eric Vanden-Eijnden\textsuperscript{2,3}}
\address{\textsuperscript{2}Machine Learning Lab, Capital Fund Management, Paris, France}
\address{\textsuperscript{3}Courant Institute, New York University, NY, USA}
\email{eve2@nyu.edu}
\date{}
\begin{document}
\setcounter{tocdepth}{1}

\vspace{-2em}
\begin{abstract}
Flow-based generative models can face numerical challenges on scientific data with multiscale Fourier spectra, often producing large errors at fine scales. We approach this problem within the flow matching and stochastic interpolants framework, through the principled design of noise distributions and interpolation schedules. Working in function space ensures that the generative model remains well defined as the resolution is refined; the Lipschitz regularity of the drift is important to both this function-space well-posedness and the integration cost at fixed resolution. The central observation is that the noise should be at least as rough as the target distribution---measured by Fourier-spectrum decay---in order to keep the Lipschitz constant finite. For Gaussian and near-Gaussian targets whose fine-scale structure is known, matched-spectrum noise improves numerical efficiency over standard white-noise choices. For more complex non-Gaussian targets, matched-spectrum noise may not be sufficient, and we propose scale-adaptive interpolation schedules to mitigate the terminal-time stiffness that arises when the noise is rougher than the data. Numerical experiments on synthetic Gaussian random fields and on invariant measures of the stochastic Allen--Cahn and Navier--Stokes equations illustrate the approach and demonstrate its ability to generate high-fidelity samples at lower computational cost than traditional approaches.
\end{abstract}
\maketitle
\vspace{-1em}
\tableofcontents

\section{Introduction}
\subsection{Context} 

Transport-based methods between probability measures using flows and diffusion processes governed by ordinary and stochastic differential equations (ODEs and SDEs) have led to remarkable successes in generative modeling across diverse domains. In computer vision, these methods have achieved state-of-the-art results in image synthesis \cite{ho2020denoising,song2020score,dhariwal2021diffusion}, super-resolution \cite{saharia2022image}, and video generation \cite{ho2022video}. Recent breakthrough applications extend to protein structure prediction \cite{abramson2024accurate}, drug discovery \cite{schneuing2022structure}, materials design \cite{zeni2025generative}, and weather forecasting \cite{price2025probabilistic}. The methodological foundations underlying these successes include score-based diffusion models \cite{song2019generative,song2020score}, flow matching \cite{lipman2022flow}, rectified flows \cite{liu2022flow}, and stochastic interpolants \cite{albergo2023building,albergo2023stochastic}.

\begin{figure}[ht]
    \centering
    \includegraphics[width=0.32\linewidth]{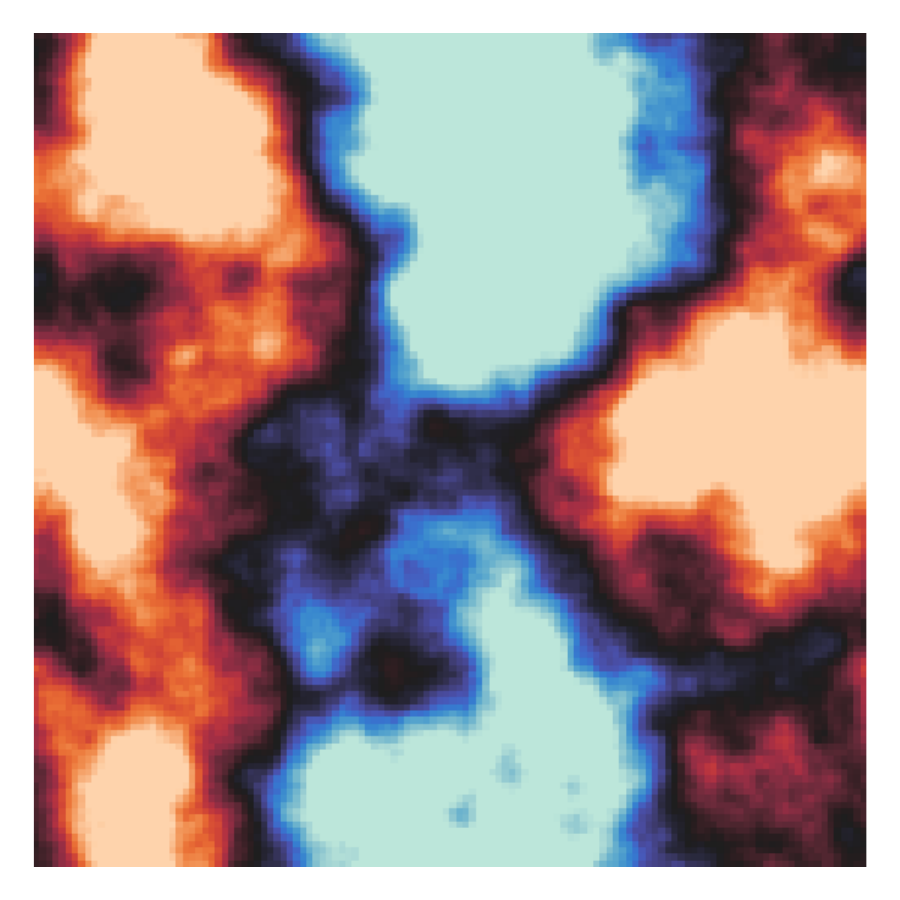}
    \includegraphics[width=0.32\linewidth]{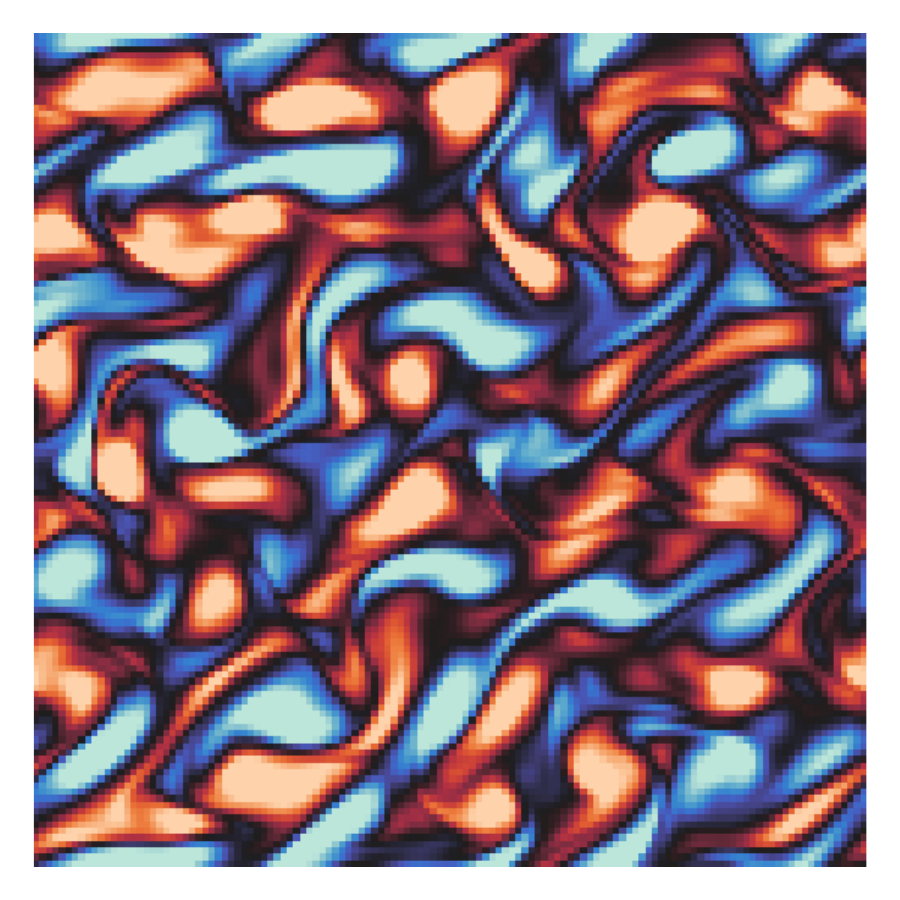}
    \includegraphics[width=0.32\linewidth]{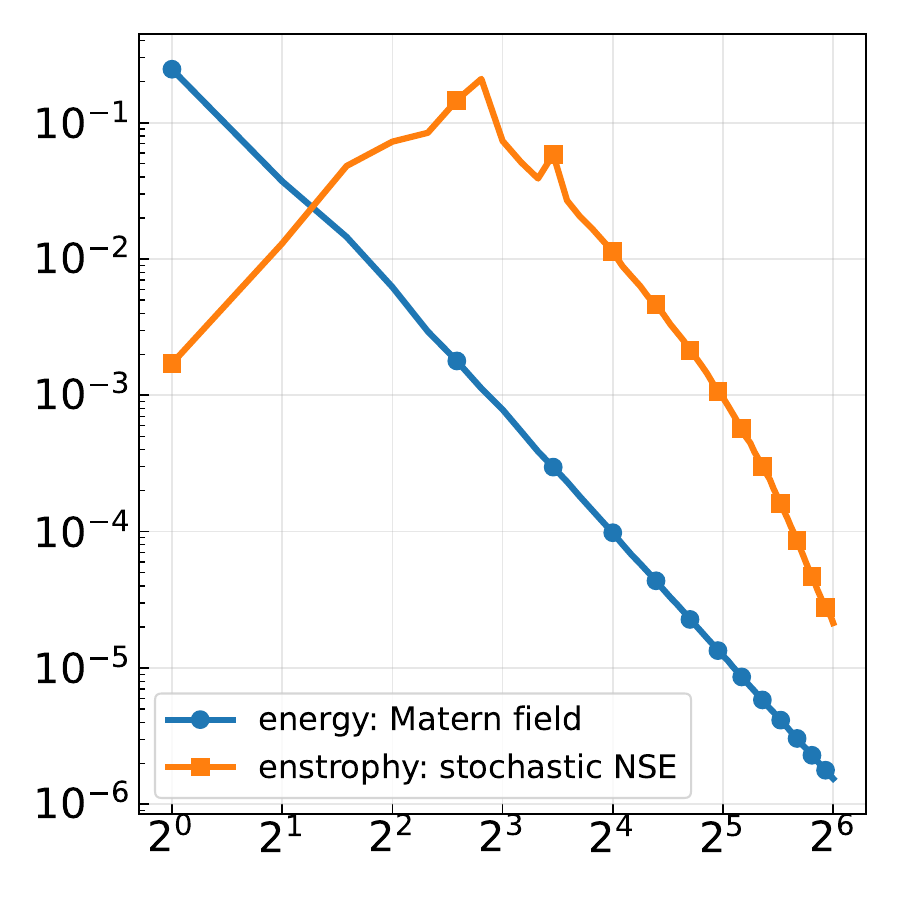}
    \caption{Examples of samples from 2D Mat\'ern Gaussian measures (left panel) and the invariant measure of the stochastically forced Navier--Stokes (middle panel) at a resolution of $128\times 128$; the right panel shows their energy and enstrophy spectra.}
    \label{fig:numerical-ill-conditioning-distributions}
\end{figure}

This paper is concerned with the application of these techniques to scientific and engineering data involving fields that exhibit numerical ill-conditioning and multiscale Fourier spectra. Such distributions present unique challenges: their Fourier spectra span multiple decades in magnitude, making accurate reproduction of fine-scale features critical yet numerically demanding. Figure \ref{fig:numerical-ill-conditioning-distributions} illustrates representative examples showing samples from a Gaussian random field and the invariant distribution of the stochastically forced Navier--Stokes equation, where spectral magnitudes vary across wide ranges of scales. Standard generative modeling approaches applied to such data often suffer from systematic errors, particularly in fine-scale spectral components that are important for physical fidelity.

We develop strategies to address these numerical challenges within the flow matching, rectified flow, and stochastic interpolants framework~\cite{lipman2022flow,albergo2023building,albergo2023stochastic,liu2022flow}, which generalizes diffusion and score-based generative models~\cite{sohl2015deep,song2019generative,ho2020denoising,song2020score} as special cases. The approach interpolates between a noise sample and a data sample, generating new samples by integrating a learned drift field. Our analysis reveals fundamental relationships between noise characteristics, interpolation schedules, and the regularity of the drift field, and we design recipes that improve efficiency and spectral accuracy when generating samples from numerically ill-conditioned distributions.

\subsection{This work}
Our contributions are as follows:

\begin{itemize}
\item \textbf{Noise selection and drift regularity} (Sections~\ref{sec-Preliminaries and Motivating Examples} and~\ref{sec-Choices of Noise for Gaussian and General Target Measures}): We analyze how the choice of noise affects the regularity of the drift. For Gaussian targets we derive a closed-form drift and show that the noise must be at least as rough as the data---in terms of Fourier-spectrum decay---for the drift to remain a bounded operator near the initial time. For non-Gaussian targets we use Cameron--Martin space theory and derive Lipschitz estimates of the drift under some assumptions. In both cases, the assumptions imply that the noise is at least as rough as the data, and the sampling ODE is shown to be well-posed in function space.

\item \textbf{Numerical efficiency and matched-spectrum noise for known structure} (Section~\ref{sec-Numerical Efficiency and Design of Noise}): We observe that rougher-than-data noise can introduce numerical stiffness near the terminal time. For distributions whose fine-scale behavior is analytically tractable---such as Gaussian random fields and stochastic Allen--Cahn invariant measures absolutely continuous with respect to a known Gaussian---we show that noise matching the data's spectrum, thus roughness, can substantially improve numerical efficiency across resolutions.

\item \textbf{Scale-adaptive schedules for unknown structure} (Section~\ref{sec-Numerical Efficiency through Design of Interpolation Schedules}): For complex non-Gaussian distributions where fine-scale structure cannot be prescribed easily \emph{a priori}---e.g., the invariant measure of stochastically forced Navier--Stokes---we observe that the matched-spectrum recipe may fail, and rougher noise is needed. We propose scale-adaptive interpolation schedules that help maintain numerical stability when using rougher-than-data noise. The resulting recipe yields more accurate enstrophy spectra together with non-Gaussian statistics such as flatness and gradient kurtosis. A simple non-Gaussianity-based diagnostic is also proposed to indicate which of the two recipes (matched noise vs.\ rougher noise plus scale-adaptive schedule) is more appropriate.
\end{itemize}

\subsection{Related work}

\subsubsection{Flows and diffusions for generative modeling}
Recent advances in generative modeling have been driven by the introduction of flow and diffusion processes that can be learned from data and efficiently integrated for sampling; see, e.g.,~\cite{yang2023diffusion} for a comprehensive review. These methods generate samples through iterative refinement processes that progressively eliminate noise or corruption across multiple scales. Critical to their performance is the design of noise distributions and scheduling strategies, which significantly impact both learning efficiency and sampling quality \cite{san2021noise,jolicoeur2021gotta,nichol2021improved,song2020improved,karras2022elucidating,shaul2024bespoke}.
Our work specifically targets data characterized by wide-range Fourier spectra, where fine-scale features require careful treatment. Beyond advanced numerical integration schemes, existing approaches for handling multiscale structure fall into two primary categories: function space generative models and multiscale hierarchical methods, which we review below.

\subsubsection{Generative models in function space}
As spatial resolution increases, data distributions are naturally viewed as measures on function spaces, and generative models defined directly in this infinite-dimensional setting aim for behavior that remains stable as the grid is refined. The same perspective has been productive in Bayesian inverse problems~\cite{stuart2010inverse} and operator learning~\cite{li2020fourier,lu2021learning}. A growing body of work pursues this direction for diffusion- and flow-based generative models~\cite{phillips2022spectral,kerrigan2023diffusion,lim2023score,pidstrigach2023infinite,bond2023infty,hagemann2023multilevel,kerrigan2023functional,baldassari2024conditional,wang2025fundiff}; see~\cite{franzese2025generative} for a survey.

These works share a common technical concern: defining generative dynamics directly on function space --- a reverse SDE for diffusion-based constructions~\cite{pidstrigach2023infinite,lim2023score,hagemann2023multilevel} or a generative ODE for flow matching~\cite{kerrigan2023functional} --- together with the corresponding score (or drift). The assumptions and emphases differ. \cite{pidstrigach2023infinite} defines the score directly as a conditional expectation in an abstract Wiener space setting and proves well-posedness of the reverse SDE in two regimes: data with bounded Cameron--Martin support, and data absolutely continuous with respect to a Gaussian reference (under a Lipschitz log-density gradient); they further give a dimension-independent Wasserstein-$2$ error bound. \cite{lim2023score} use trace-class Gaussian-process forward noise, define the score by generalizing denoising score matching to function space (approximated with neural operators), and sample with a function-valued annealed Langevin dynamic whose well-posedness is invoked from prior infinite-dimensional Langevin results. \cite{hagemann2023multilevel} trains diffusion models simultaneously on multiple discretization levels of infinite-dimensional functions and shows that the multilevel approach is consistent in the infinite-dimensional limit. Functional flow matching (FFM)~\cite{kerrigan2023functional} extends flow matching to function space with the conditional flow construction using a trace-class Gaussian noise; it establishes marginal-matching under regularity and absolute-continuity assumptions on the conditional construction, but does not verify those assumptions for concrete noise--data pairs.

We work with generative ODEs, and the method is most closely related to FFM. Working directly with the stochastic interpolant construction, we establish well-posedness of the sampling ODE in function space (Propositions~\ref{prop-gaussian-measure}, \ref{prop-lip-CM-space}, and~\ref{prop-well-posedness}), for Gaussian measure targets as well as the setup of~\cite{pidstrigach2023infinite} for the Cameron--Martin-supported case. We then additionally study two design degrees of freedom --- the noise covariance $C_0$ (Section~\ref{sec-Numerical Efficiency and Design of Noise}) and the time schedule $(\alpha_t,\beta_t)$ (Section~\ref{sec-Numerical Efficiency through Design of Interpolation Schedules}), to improve numerical performance. We note that function-space well-posedness can ensure stable coarse-scale behavior under resolution refinement; that is how Wasserstein-$2$ distances remain stable and bounded as resolution increases, since $W_2$ averages over scales. We pay attention to precise reproduction of the fine-scale spectral features that are important for physical fidelity, at a finite resolution.

\subsubsection{Multiscale generative models}
Multiscale and hierarchical architectures have improved image diffusion models through progressive refinement strategies~\cite{dhariwal2021diffusion,jing2022subspace,saharia2022image,ho2022cascaded,rissanen2022generative,hoogeboom2023blurring}. Complementary approaches leverage wavelet decompositions to exploit natural multiscale structure~\cite{yu2020wavelet,marchand2022wavelet,guth2022wavelet,phung2023wavelet,lempereur2024hierarchic}. These methods connect to renormalization-group theory, where flows naturally exhibit fine-to-coarse directional structure~\cite{cotler2023renormalization,cotler2023renormalizing,sheshmani2024renormalization}. 
 Building on similar principles, we demonstrate that scale-adaptive design
of noise distributions and interpolation schedules can substantially improve numerical performance while maintaining accurate reproduction across all scales with modest
computational overhead.

\section{Preliminaries, Motivating Examples, and Function Space}
\label{sec-Preliminaries and Motivating Examples}
\subsection{Flow matching and stochastic interpolants on $\mathbb R^d$}
\label{sec-SI-Rd}
We work within the flow matching and stochastic interpolants framework~\cite{lipman2022flow,albergo2023building,albergo2023stochastic,liu2022flow}. For clarity of exposition we first review the construction on Euclidean $\mathbb R^d$; proof sketches are in Appendix~\ref{appendix:derivation-stochastic-interpolants}. The function-space extension is presented in Section~\ref{sec-SI-function-space}. In this paper we do not explicitly distinguish the a.e.-defined $L^2$ element and its pointwise defined version for simplicity of presentation; the latter should be understood as a particular regular version of the former.

\begin{definition}
    Given a target distribution $\mu^*$ satisfying $\int_{\mathbb{R}^d} \|x\|^2 \mu^*({\rm d}x) < \infty$, the linear stochastic interpolant between $x_1 \sim \mu^*$ and the independent Gaussian noise $z \sim \sfN(0,\mathrm{I})$ with $z \perp x_1$ is defined as
    \begin{equation}
    \label{eqn-linear-interpolant}
         I_t = \alpha_t z + \beta_t x_1, \quad 0 \leq t \leq 1\, .
    \end{equation}
    Here $\alpha_t, \beta_t \in C^1([0,1])$ are scalar interpolation schedules satisfying the boundary conditions $\alpha_0 = \beta_1 = 1$ and $\alpha_1 = \beta_0 = 0$. We also assume $\dot{\beta}_t > 0$ and $\dot{\alpha}_t < 0$ throughout this paper.
\end{definition}

The noise can be taken as an arbitrary distribution; the Gaussian choice above is convenient. Given the interpolant, we can identify the path of marginals $\mu_t = \mathrm{Law}(I_t)$ with the law of an ODE flow generated by a conditional-expectation drift:

\begin{proposition}
\label{prop-si-bt}
The conditional expectation $b_t(x) := \mathbb{E}[\dot I_t \mid I_t = x]$ is well defined for every $t\in[0,1]$ as an element of $L^2(\mathrm d\mu_t;\mathbb R^d)$. Assume the ODE $\dot X_t = b_t(X_t)$ has a unique strong solution for $\sfN(0,\mathrm I)$-a.e.\ initial condition, and that, with $X_0\sim\sfN(0,\mathrm I)$, the curve of marginals $\nu_t := \mathrm{Law}(X_t)$ is the unique weak solution of the continuity equation $\partial_t\nu_t + \nabla\!\cdot\!(b_t\,\nu_t) = 0$ with $\nu_0 = \sfN(0,\mathrm I)$. Then $\nu_t = \mu_t$ for every $t\in[0,1]$; in particular $X_1\sim\mu^*$.
\end{proposition}

A proof is in Appendix~\ref{appendix:derivation-stochastic-interpolants}. The proposition implies that when $b_t$ is regular enough that both the ODE and its associated continuity equation are well-posed, the ODE flow generates the interpolant marginals. This is the basis for the numerical algorithm: sample $X_0\sim\sfN(0,\mathrm I)$, integrate the ODE, output $X_1\sim\mu^*$.

Since $b_t$ is a conditional expectation, it can be estimated from samples by minimizing the squared loss
\begin{equation}
\label{eqn-loss-Rd}
    L(\hat b) = \int_0^1 \mathbb{E}\big[\|\hat b_t(I_t) - \dot I_t\|_2^2\big]\,{\rm d}t,
\end{equation}
parametrizing $\hat b$ by a neural network. The learned drift $\hat b$ is then integrated by a numerical scheme. Alternative stochastic differential equation (SDE) formulations are also possible \cite{albergo2023stochastic}; we focus on the ODE formulation throughout.

\subsection{Motivating example: the choice of noise matters}
\label{sec-Motivating example: the choice of noise matters}
When $x_1$ is high-dimensional, for example arising from discretized continuous fields in physical sciences, the choice of noise distribution becomes important, as we illustrate with the concrete example below. 
\begin{example}
Consider a one-dimensional domain $D = [0,1]$ and a spatial Gaussian process $\xi \sim \mathcal{GP}(0,k)$ with covariance kernel $k(y,z) = \exp(-\|y-z\|_2^2/(2l^2))$ for lengthscale $l=1$. Let $x_1 \in \mathbb{R}^{N}$ be a discretization on a uniform grid with spacing $h=1/N$. By construction, $x_1 \sim \sfN(0, K_{N})$ where $K_{N} \in \mathbb{R}^{N \times N}$ is the covariance matrix with diagonal entries equal to one.
As $N\to \infty$, the eigenvalues of $\frac{1}{N}K_{N}$ converge to those of the integral operator 
\[(\mathcal{K}f)(y) = \int_D k(y,z) f(z)\, {\rm d}z\, . \]

A natural choice for noise is $z \sim \sfN(0,\mathrm{I}_{N})$, which preserves variance since $\mathbb{E}[\|z\|^2_2] = N = \mathbb{E}[\|x_1\|^2_2]$. Here we write the subscript $N$ explicitly. However, with this choice for $z$, if $x_1 \sim \sfN(0,K_{N})$ and $\alpha_t, \beta_t \in C^1([0,1])$, then
\begin{equation}
    b_t(x) = \mathbb{E}[\dot{I}_t  | I_t = x] = B_{N}(t) x\, ,
\end{equation}
where 
\[B_{N}(t) = (\dot{\alpha}_t \alpha_t I_{N} + \dot{\beta}_t \beta_t K_{N})(\alpha_t^2 I_{N} + \beta_t^2 K_{N})^{-1}\, .\] 
As a result, $\lim_{t \to 0} \lim_{N\to \infty} \|B_N(t)\|_2 = \infty$, and the drift $b_t(x)$ also diverges in this limit.

This problem arises because of the scale imbalance between $\mathrm{I}_N$ (spectral norm $O(1)$) and $K_N$ (spectral norm $O(N)$), despite having equal traces. As $N \to \infty$, the noise $\sfN(0,\mathrm{I}_N)$ becomes trivial while $x_1$ converges to a non-trivial Gaussian process.

To have a balanced design, one should instead use the noise $z \sim \sfN(0, N \mathrm{I}_{N})$ at $N$ grid points; then this process will converge to a non-trivial white noise in the continuous limit $N \to \infty.$
\end{example}

The above motivating example shows that for distributions arising from continuous fields, the noise has to be designed so that the model remains meaningful as the number of grid points grows. This observation---that a generative model for function-valued data should be posed directly in the infinite-dimensional setting---underlies the function-space generative-modeling literature~\cite{pidstrigach2023infinite,lim2023score,kerrigan2023functional,hagemann2023multilevel}. We develop our function-space setting in the next subsection.

\subsection{The function-space framework}
\label{sec-SI-function-space}

Let $H$ be a separable Hilbert space and let $z\sim\sfN(0,C_0)$ be an $H$-valued Gaussian random variable (equivalently, its covariance $C_0$ is trace-class on $H$), and let $x_1\sim\mu^*$ be independent of $z$ with $\int_H \|x\|_H^2\,\mu^*({\rm d}x)<\infty$. The interpolant $I_t = \alpha_t z + \beta_t x_1$, $t\in[0,1]$, uses the same schedules as in Section~\ref{sec-SI-Rd}; we write $\mu_t = \mathrm{Law}(I_t)$. Probability densities are not well defined on $H$ in general, but conditional expectations are, so we use them in the construction. The proof can be found in Appendix~\ref{appendix:derivation-stochastic-interpolants}.

\begin{proposition}
\label{prop-si-bt-functionspace}
The conditional expectation $b_t(x) := \mathbb{E}[\dot I_t \mid I_t = x]$ is well defined for every $t\in[0,1]$ as an element of $L^2(\mathrm d\mu_t;H)$. Assume the ODE $\dot X_t = b_t(X_t)$ has a unique strong solution for $\sfN(0,C_0)$-a.e.\ initial condition, and that, with $X_0\sim\sfN(0,C_0)$, the curve of marginals $\nu_t := \mathrm{Law}(X_t)$ is the unique weak solution of the continuity equation $\partial_t\nu_t + \nabla\!\cdot\!(b_t\,\nu_t) = 0$ with $\nu_0 = \sfN(0,C_0)$. Then $\nu_t = \mu_t$ for every $t\in[0,1]$; in particular $X_1\sim\mu^*$.
\end{proposition}

As in the Euclidean case, the bundled hypothesis will hold under sufficient regularity of the drift $b_t$. Section~\ref{sec-Choices of Noise for Gaussian and General Target Measures} establishes this regularity in two settings of interest and bundles it with the marginal-matching identity above into a single well-posedness statement (Proposition~\ref{prop-well-posedness}). 

The above is similar to the framework of functional flow matching \cite{kerrigan2023functional}, which adopts the flow-matching construction using conditional flow and velocity fields, while we use a direct and explicit interpolant construction. They are complementary perspectives, with somewhat different technical assumptions and arguments to make them rigorous in the function-space setting. 

One can estimate $b_t$ by minimizing the $L^2(\mathrm d\mu_t;H)$ loss
\begin{equation}
\label{eqn-loss-fs}
    L(\hat b) = \int_0^1 \mathbb{E}\big[\|\hat b_t(I_t) - \dot I_t\|_H^2\big]\,\mathrm{d}t,
\end{equation}
which extends~\eqref{eqn-loss-Rd} to the function-space setting; for fields on an $N\times N$ grid we can approximate $\|\cdot\|_H$ by the per-pixel uniformly weighted $\ell^2$ norm. In general, one can replace the $H$-norm with a higher-regularity norm to penalize the high-frequency parts more strongly; see for example \cite{pidstrigach2023infinite}. We do not focus our analysis on the loss-norm choice here; we train sufficiently long in all our experiments, and the plain $L^2$ loss was sufficient in our experiments (using a weighted norm does not change the results). A finer analysis of how the loss-norm choice interacts with spectral sensitivity is a direction we leave for future work.

\section{Drift Regularity: Gaussian and General Target Measures}
\label{sec-Choices of Noise for Gaussian and General Target Measures}
We now study the regularity of the drift $b_t$ for two classes of target measures. Our aim is to derive Lipschitz regularity of the drift field and to establish well-posedness of the sampling ODE in function space. Section~\ref{sec-noise-for-gaussian} treats Gaussian targets using the explicit conditioning formula, and Section~\ref{sec-noise-for-general} treats general targets via Cameron--Martin space theory and closes with the well-posedness statement that consolidates both cases.

\subsection{Gaussian target measures}
\label{sec-noise-for-gaussian}
We specialize to a Gaussian target: $x_1\sim\sfN(0,C_1)$ is an $H$-valued Gaussian random variable, with covariance $C_1$ a positive, self-adjoint, trace-class operator on $H$~\cite{bogachev1998gaussian}. When both $z\sim\sfN(0,C_0)$ and $x_1$ are Gaussian, $I_t$ is Gaussian as well and the drift admits a closed form.
\begin{proposition}
\label{prop-gaussian-measure}
Suppose $z\sim\sfN(0,C_0)$ and $x_1\sim\sfN(0,C_1)$ are independent Gaussian measures on $H$, with $C_0,C_1$ simultaneously diagonalizable: there is an orthonormal basis $\{e_j\}_{j\ge1}$ of $H$ with $C_0 e_j = c_0(j)\,e_j$ and $C_1 e_j = c_1(j)\,e_j$, where $c_0(j),c_1(j)>0$. Then $I_t\sim\mu_t := \sfN(0,\Sigma_t)$ with $\Sigma_t := \alpha_t^2 C_0 + \beta_t^2 C_1$ a positive, self-adjoint, trace-class operator on $H$. For every $t\in(0,1)$ the drift $b_t(x) := \bE[\dot I_t \mid I_t = x]$ is defined for $\mu_t$-a.e.\ $x\in H$ as the regular conditional expectation of the jointly Gaussian pair $(\dot I_t, I_t)$, which is a measurable linear function of $x$, namely $b_t(x) = B(t)x$. On the dense (but $\mu_t$-null) subspace $\mathrm{Ran}(\Sigma_t)$ it is given by
\[
B(t) := (\dot\alpha_t \alpha_t C_0 + \dot\beta_t \beta_t C_1)\,\Sigma_t^{-1}.
\]
In the basis $\{e_j\}$ the operator $B(t)$ is diagonal with multipliers
\[
b_j(t) = \frac{\dot\alpha_t \alpha_t\, c_0(j) + \dot\beta_t \beta_t\, c_1(j)}{\alpha_t^2\, c_0(j) + \beta_t^2\, c_1(j)},
\]
so it extends to a bounded operator on $H$ with $\|B(t)\| = \sup_j |b_j(t)|$. Consequently, if $\sup_j c_1(j)/c_0(j) < \infty$ (equivalently, $C_1 C_0^{-1}$ admits a bounded extension to $H$), then $\|B(t)\|$ remains bounded as $t\to 0$; if $\sup_j c_1(j)/c_0(j) = \infty$, then $\|B(t)\|\to\infty$ as $t\to 0$.
\end{proposition}
The proof is in Appendix~\ref{sec-Proof for Gaussian Target Measures}. The takeaway is that boundedness of $B(t)$ near the initial time is governed by the relative spectral decay of $C_1$ and $C_0$: if $C_1$ decays no slower than $C_0$ (i.e., $C_1 C_0^{-1}$ admits a bounded extension to $H$), the drift is a bounded operator $H\to H$ uniformly near $t=0$. When this fails, $\|B(t)\|$ diverges as $t\to 0$; formal solutions may still exist mode by mode, but the ODE is ill-conditioned for numerical integration, with small perturbations at the initial time amplified without bound at fine scales.

\begin{remark}[Trace-class is relative to $H$; white noise is included]
\label{rmk-trace-class-relative}
The trace-class hypothesis is relative to the ambient $H$ and does not restrict the roughness of the noise; it only fixes the space on which $\sfN(0,C_0)$ is a genuine Gaussian measure. White noise is the prime example: its covariance is not trace-class on $L^2$, but on a negative-Sobolev space $H = H^{-r}(D)$ with $r > d/2$ it is, so white noise is a bona fide Gaussian measure on $H$, while the smoother data still lives in $H$ since $L^2\subset H^{-r}$. Enlarging $H$ to host the noise thus costs nothing. The substantive condition is instead the \emph{relative} spectral decay: $B(t)$ stays bounded near $t=0$ precisely when $C_1 C_0^{-1}$ admits a bounded extension to $H$, i.e.\ the noise is at least as rough as the data---a property of the noise--data pair.
\end{remark}

\subsubsection{Example: Mat\'ern fields}
\label{sec-Example: Matern fields}
Let us examine the above condition for the specific example of Mat\'ern-like fields, which are
commonly used Gaussian process (GP) models in spatial statistics.
Consider a Mat\'ern-like Gaussian measure $\xi \sim \sfN(0,\, \sigma^2(-\Delta + \tau^2 \mathrm{I})^{-s})$
on the two-dimensional torus $\mathbb{T}^2 = [0,1]^2$ (i.e., $D=[0,1]^2$ with periodic boundary
conditions). On $\mathbb{T}^2$ the negative Laplacian $-\Delta$ is diagonalized by the Fourier basis
$e_m(y) = e^{2\pi i\langle m, y\rangle}$, $m\in\mathbb{Z}^2$, with eigenvalues
$\lambda_m = 4\pi^2|m|^2$. A sample can therefore be drawn mode by mode through its Fourier
coefficients,
\begin{equation*}
    \widehat{\xi}(m) = \sigma\,(\lambda_m + \tau^2)^{-s/2}\,\xi_m, \qquad m\in\mathbb{Z}^2\setminus\{0\}\, ,
\end{equation*}
where the $\xi_m$ are independent standard complex Gaussians subject to the reality constraint
$\xi_{-m} = \overline{\xi_m}$, and the constant mode $m=0$ is omitted. (Equivalently, in a real
trigonometric basis, $\xi(y)=\sum_m \sigma(\lambda_m+\tau^2)^{-s/2}\phi_m(y)\,\xi_m$ with $\{\phi_m\}$
the orthonormal cosine/sine modes and $\xi_m$ independent standard normals.)

Let $C_0 = \sigma_0^2(-\Delta + \tau_0^2 \mathrm{I})^{-s_0}$ and
$C_1 = \sigma_1^2(-\Delta + \tau_1^2 \mathrm{I})^{-s_1}$. Both are diagonal in this Fourier basis, with
eigenvalues $c_i(m) = \sigma_i^2(4\pi^2|m|^2 + \tau_i^2)^{-s_i}$; hence $C_1 C_0^{-1}$ is diagonal in this basis with eigenvalues
$c_1(m)/c_0(m)$ and admits a bounded extension to $H$ if and only if $s_0\leq s_1$. This means the noise process
$\sfN(0,C_0)$ should be rougher than, or at least as rough as, the data $\sfN(0,C_1)$, which is
necessary for the generative ODE to have a drift that is a bounded operator near the initial time.

\begin{remark}
\label{connect to whittle matern}
The parameters $\sigma$, $\tau$, and $s$ characterize the process amplitude, inverse lengthscale, and regularity, respectively. This parameterization parallels the standard Mat\'ern process \cite{stein1999interpolation,guttorp2006studies} defined on $\mathbb{R}^d$, whose kernel function and covariance operator are similarly determined by three parameters. The connection to solutions of stochastic PDEs, pioneered by Whittle \cite{whittle1954stationary,guttorp2006studies}, is explored in \cite{lindgren2011explicit}. The Mat\'ern kernel function is expressed as:
\[K_{\sigma,l,\nu}(x,y)=\sigma^2\frac{2^{1-\nu}}{\Gamma(\nu)}\left(\frac{|x-y|}{l}\right)^\nu B_\nu\left(\frac{|x-y|}{l}\right)\, , \]
for $x,y \in \mathbb{R}^d$, where $B_\nu$ represents the modified Bessel function of the second kind of order $\nu$. On $\mathbb{R}^d$, this kernel corresponds to the covariance operator:
\[C_{\sigma,l,\nu} = \frac{\sigma^2l^d\Gamma(\nu+d/2)(4\pi)^{d/2}}{\Gamma(\nu)}(I-l^2\Delta)^{-\nu-d/2}\, .\]
This formulation illuminates the relationship between the Mat\'ern covariance operator on $\mathbb{R}^d$ and our Mat\'ern-like covariance operator defined on the bounded domain. We focus our examples on the bounded domain to leverage Fourier-series techniques.
\end{remark}

\subsection{General target measures via Cameron--Martin space}
\label{sec-noise-for-general}
For non-Gaussian targets the drift has no closed form. The Lipschitz regularity can be studied by working with the Cameron--Martin space of the noise. Recall~\cite{bogachev1998gaussian,hairer2009introduction} that the Cameron--Martin space of the Gaussian noise $\sfN(0,C_0)$ is the Hilbert space $V = C_0^{1/2}(H)$ with $\langle y, z\rangle_V = \langle C_0^{-1/2} y, C_0^{-1/2} z\rangle_H$. The Gaussian noise lives in $H\setminus V$ almost surely, so $\sfN(0,C_0)$ is canonically rougher than a measure supported on $V$; for example, Brownian motion lives in $H = L^2$ but not in $V = H^1$. We state the Lipschitz result under the assumption that the target has bounded $V$-support (i.e.\ $\|x_1\|_V\le R$ almost surely), a standard simplification in theoretical analyses~\cite{chen2023sampling,gao2023gaussian,pidstrigach2023infinite}.
\begin{proposition}
\label{prop-lip-CM-space}
Assume $x_1\sim\mu^*$ is supported in $V$ with $\|x_1\|_V\le R$ a.s., and let $m_t(x) := \mathbb{E}[x_1\mid I_t = x]$, which is called the \emph{denoiser}.
\begin{itemize}
\item[(i)] \emph{Boundedness of the denoiser in $V$.} For every $t\in[0,1)$, $m_t$ is a map $H\to V$ with $\|m_t(x)\|_V\le R$ for $\mu_t$-a.e.\ $x\in H$. The drift admits the representation
\begin{equation}
\label{eqn-bt-denoiser-form}
b_t(x) = \tfrac{\dot\alpha_t}{\alpha_t}\,(x-\beta_t m_t(x)) + \dot\beta_t\, m_t(x).
\end{equation}
\item[(ii)] \emph{Lipschitz continuity ($V\to V$).} For every $\delta\in(0,1)$, $m_t\colon V\to V$ is Lipschitz uniformly in $t\in[0,1-\delta]$, with Lipschitz constant $\tilde L_{1-\delta}$ depending only on $\delta$, $R$, and the schedule.
\end{itemize}
\end{proposition}
The proof is in Appendix~\ref{sec-Proof for General Target Measures}; the argument follows the bounded-Cameron--Martin-support setting of~\cite[Theorem~12]{pidstrigach2023infinite}. The assumption $x_1\in V$ a.s.\ implies a rougher-than-data condition, since $\sfN(0,C_0)$ is almost surely \emph{not} in $V$. 

We close this section by presenting the well-posedness of the sampling ODE on $H$, which combines the regularity results of Propositions~\ref{prop-gaussian-measure} and~\ref{prop-lip-CM-space} with Proposition~\ref{prop-si-bt-functionspace}.

\begin{proposition}[Well-posedness of the sampling ODE on $H$]
\label{prop-well-posedness}
Fix $\delta\in(0,1)$ and consider the sampling ODE $\dot X_t = b_t(X_t)$, $X_0\sim\sfN(0,C_0)$, on $H$. Assume either
\begin{itemize}
\item[(a)] the Gaussian setting (Proposition~\ref{prop-gaussian-measure}) with $C_1 C_0^{-1}$ bounded on $H$; or
\item[(b)] the bounded-$V$-support setting (Proposition~\ref{prop-lip-CM-space}) for more general $\mu^*$.
\end{itemize}
Then, for $\sfN(0,C_0)$-a.e.\ initial condition $X_0$, the ODE admits a unique strong solution $X_t\in H$ on $[0,1-\delta]$, and $\mathrm{Law}(X_t)=\mu_t$ for every $t\in[0,1-\delta]$. 
\end{proposition}

\smallskip
In case (a), $b_t(x) = B(t)\,x$ is linear with $B(t)\colon H\to H$ uniformly bounded on $[0,1-\delta]$, so Picard--Lindel\"of on $H$ applies directly. In case (b), the drift has the \emph{semilinear} structure
\[
b_t(x) = \tfrac{\dot\alpha_t}{\alpha_t}\,x + \big(\dot\beta_t - \tfrac{\beta_t\dot\alpha_t}{\alpha_t}\big)\,m_t(x),
\]
in which the nonlinear part $m_t$ is $V$-valued. Variation of constants then implies that the trajectory lies in the affine subspace $\alpha_t X_0 + V$, so the correction $W_t := X_t - \alpha_t X_0$ takes values in $V$ and satisfies an ODE whose drift is $V\to V$ Lipschitz. Picard--Lindel\"of on $V$ then delivers a unique $W_t$, and $X_t = W_t + \alpha_t X_0$ is the unique $H$-valued solution. Marginal-matching follows from Proposition~\ref{prop-si-bt-functionspace} in both cases. Full details are in Appendix~\ref{sec-Proof for General Target Measures}.

The terminal time $t=1$ is often delicate and is handled in practice ~\cite{karras2022elucidating} by stopping at $t=1-\delta$ for a small $\delta>0$ (we use $\delta = 10^{-3}$, see Appendix~\ref{appendix-reproducibility}).

\section{Numerical Efficiency and Design of Noise}
\label{sec-Numerical Efficiency and Design of Noise}

The discussions in the previous section show that we should choose noise that is at least as rough as the data distribution. This can guarantee that the drift is at least a bounded operator near the initial time. 

Nevertheless, the discussion does not highlight the behavior of the drift at the terminal time $t=1$. There can, in fact, be numerical issues associated with this behavior. In Section \ref{sec-The need for numerical efficiency}, we discuss this issue in the Gaussian measure setting. Section \ref{sec-Numerical efficiency through spectrum noise} considers using a specific matched-spectrum noise to handle the issue when the data's fine-scale structure is analytically tractable and uses numerical experiments to understand its strengths and limitations.

\subsection{The need for numerical efficiency}
\label{sec-The need for numerical efficiency}

Consider the Gaussian measure example in Proposition \ref{prop-gaussian-measure}. The drift is given by
\[b_t(x) = \bE[\dot I_t  | I_t = x] =  B(t) x \, ,\]
where $B(t)$ is a linear operator defined as
\[B(t) x = (\dot\alpha_t \alpha_t C_0 + \dot\beta_t \beta_t C_1)(\alpha_t^2 C_0 + \beta_t^2 C_1)^{-1}x\, . \]
When $C_1C_0^{-1}$ is bounded, the operator $B(t)$ remains bounded near $t=0$. However, as $t\to 1$, the operator becomes unbounded if $C_0C_1^{-1}$ is unbounded. The Mat\'ern-like example illustrates this behavior: we take $C_0 = \sigma_0^2(-\Delta + \tau_0^2 \mathrm{I})^{-s_0}$ and $C_1 = \sigma_1^2(-\Delta + \tau_1^2 \mathrm{I})^{-s_1}$ with $s_0 \leq s_1$, so that the noise is rougher than the data. Then, using the standard linear schedule $\alpha_t = 1 - t$ and $\beta_t = t$, we obtain 
\[B(t) = ((t-1) + t C_1C_0^{-1}) ((t-1)^2 + t^2C_1C_0^{-1})^{-1}\, . \]
In Fourier space, this operator becomes diagonal. For mode $m \in \mathbb{Z}^2_+ \backslash \{0\}$, we have 
\[\tilde{B}(t;m) = \frac{(t-1)+t \mu_m}{(t-1)^2 + t^2 \mu_m} \, ,\]
where
\[\mu_m = \frac{\sigma_1^2 (4\pi^2\|m\|^2_2 + \tau_1^2)^{-s_1} }{\sigma_0^2 (4\pi^2\|m\|^2_2 + \tau_0^2)^{-s_0}} \, .\]
The behavior depends critically on the smoothness relationship:
\begin{itemize}
    \item \textbf{Case 1: $s_0 = s_1$} (matched smoothness). Here $\mu_m$ is uniformly bounded above and below, preventing any blow-up in $\tilde{B}(t;m)$ at any time.
    \item \textbf{Case 2: $s_0 < s_1$} (rougher noise). We have $\lim_{m\to \infty} \mu_m = 0$, which yields
\[\lim_{m\to \infty} \tilde{B}(t;m) = -\frac{1}{1-t}\, .\]
\end{itemize}
For \textbf{Case 2}, as $t$ approaches 1, high-frequency modes experience unbounded sensitivity. While the negative sign prevents error amplification to infinity (unlike the case $s_0 > s_1$ near $t=0$), this increasing sensitivity to fine-scale modes necessitates very small stepsizes during ODE integration to capture the fine-scale information accurately. 

The exception is $s_0 = s_1$, where we have exact knowledge of the data's fine-scale asymptotics and can directly match the noise accordingly.

\subsection{Numerical efficiency through spectrum noise}
\label{sec-Numerical efficiency through spectrum noise}
Motivated by the discussion in the previous section, when we have precise knowledge of fine-scale behavior, choosing noise that matches the data's fine-scale structure is a natural option. This section examines the consequences for numerical efficiency through experiments.

In particular, Section \ref{sec-example-Gaussian} discusses a synthetic Gaussian measure example, and Section \ref{sec-example-stochastic Allen--Cahn} presents an example of invariant distributions of stochastic Allen--Cahn equations. Section \ref{sec-failure-example-navier-stokes} examines a challenging case involving invariant distributions of stochastically forced Navier--Stokes equations, where unfortunately noise with matching smoothness fails to achieve accurate results. The following Section \ref{sec-Numerical Efficiency through Design of Interpolation Schedules} then develops alternative interpolation schedules to overcome this limitation.

For the Gaussian target the drift admits a closed form (Proposition~\ref{prop-gaussian-measure}), which we evaluate directly via FFT; no neural network is trained. For the Allen--Cahn and Navier--Stokes targets the drift is learned with a $2$M-parameter UNet \cite{ho2020denoising}. We integrate the resulting ODE with fixed-step RK4, so the number of drift evaluations per sample is $\text{NFE}=4\times(\text{steps})$ in every case, and wall-clock cost scales linearly with NFE; in the Gaussian case each evaluation is an FFT (sub-millisecond per sample at the resolutions we consider), while in the Allen--Cahn and Navier--Stokes cases each evaluation is one UNet forward pass. The code is available at
\begin{center}\url{https://github.com/yifanc96/GenerativeDynamics-NumericalDesign.git}.
\end{center}
Full details on data generation, sample counts, training protocol, and seeds are reported in Appendix~\ref{appendix-reproducibility}.
\subsubsection{Example: Gaussian measures}
\label{sec-example-Gaussian}
As a demonstration, we consider the 2D Mat\'ern-like Gaussian measure from Section \ref{sec-Example: Matern fields}. We take noise $z \sim \sfN(0,C_0)$ and target $x_1 \sim \sfN(0, C_1)$, where $C_1 = \sigma_1^2(-\Delta + \tau_1^2 \mathrm{I})^{-s_1}$ with $s_1=3$, $\tau_1 = 1$, and $\sigma^2_1 = (4\pi^2+\tau^2_1)^{s_1}$. 

We consider two choices for the noise covariance $C_0=\sigma_0^2(-\Delta + \tau_0^2 \mathrm{I})^{-s_0}$: white noise (with $\sigma_0=1, s_0 = 0$) and matched-spectrum noise (identical to $C_1$). We term the latter \textit{spectrum noise} since it has the same Fourier spectrum as the target distribution.

For both cases, we discretize the 2D field on an $N \times N$ grid and use the interpolant $I_t = \alpha_t z + \beta_t x_1$ with linear schedule $\alpha_t = 1-t$ and $\beta_t = t$ to construct ODE generative models. Training datasets contain $5\times 10^4$ samples drawn directly from the target Gaussian, and the noise spectrum is matched analytically (no spectrum estimation needed). The ODE is solved using fourth-order Runge-Kutta (RK4) with varying numbers of integration steps.

For accuracy evaluation, we use the energy spectrum of generated samples as our criterion. For a 2D sample $u$ of side $N$ with Fourier coefficients $\hat{u}(m):=\mathcal F u$, the radially-averaged spectrum is
\[S(k) = \pi(k_+^2-k_-^2)\cdot \mathrm{mean}\{|\hat{u}(m)|^2 : \|m\|_2\in[k_-,k_+)\}\, ,\]
where $(k_-,k_+)=(k-0.5,k+0.5)$ and $k$ ranges over integer wavenumbers from $1$ to $N/2$ (in the 1D Allen--Cahn case of Section~\ref{sec-example-stochastic Allen--Cahn}, $S(k)$ is the ensemble mean of $|\hat{u}(k)|^2$ over $\pm k$). We compute $S(k)$ by averaging over a sufficiently large ensemble of samples for each frequency $k$.
\begin{figure}[ht]
    \centering
    \includegraphics[width=0.32\linewidth]{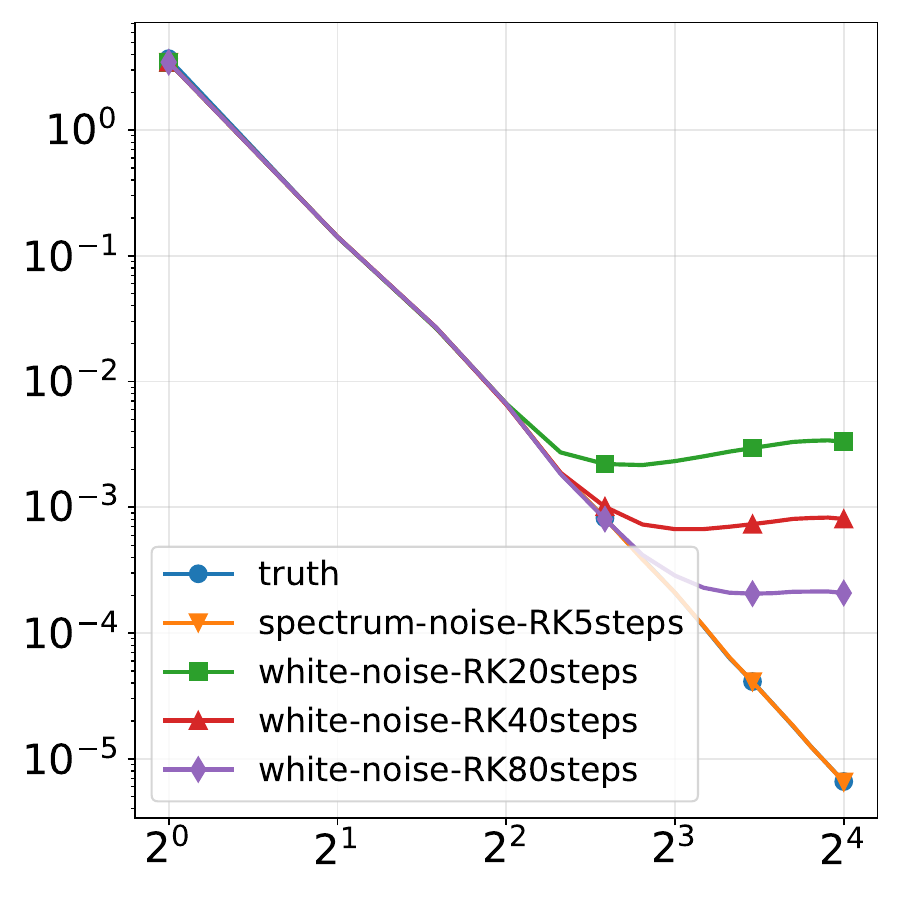}
    \includegraphics[width=0.32\linewidth]{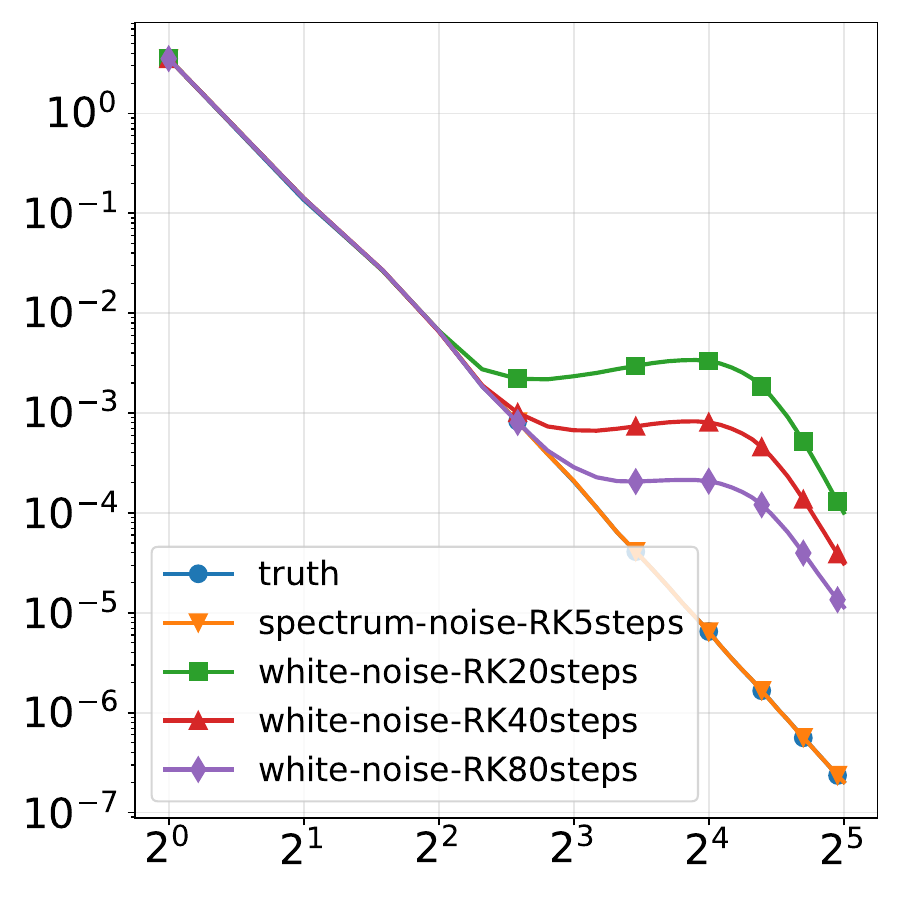}
    \includegraphics[width=0.32\linewidth]{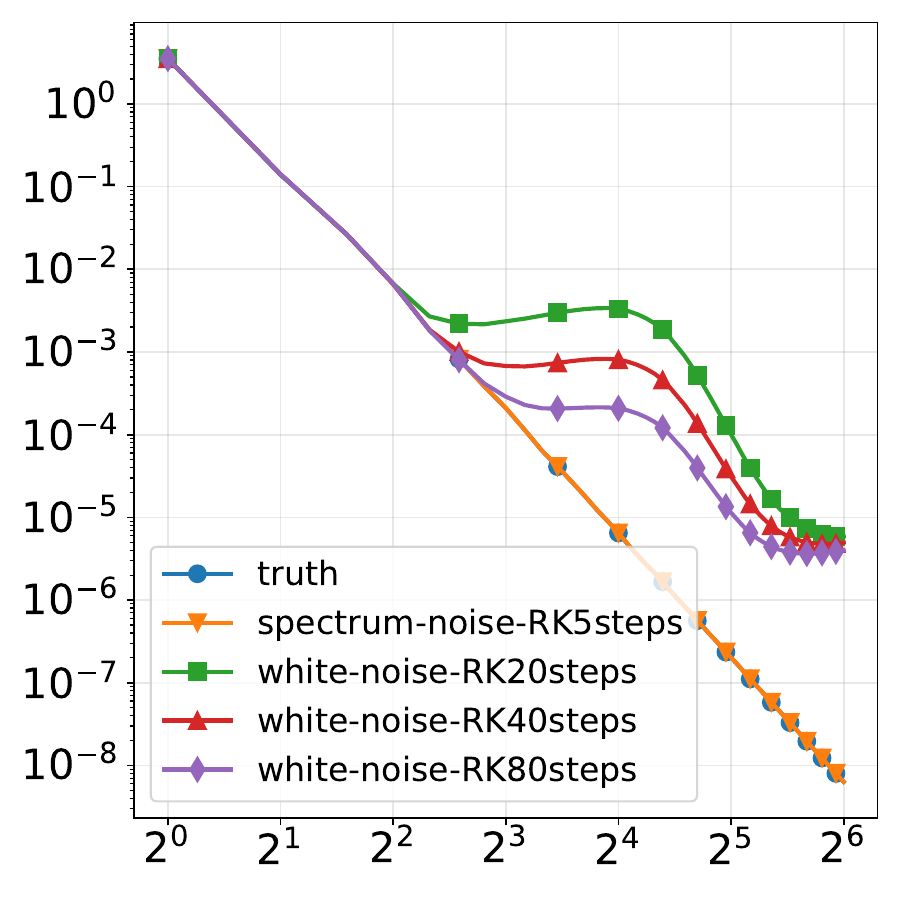}
    \caption{Energy spectra of Gaussian fields: comparison between ground truth, spectrum noise generation (5 RK4 steps), and white noise generation (20, 40, or 80 RK4 steps). Left: $32\times 32$ resolution; middle: $64\times 64$; right: $128\times 128$.}
    \label{fig:gaussian-spectrum}
\end{figure}

Figure \ref{fig:gaussian-spectrum} compares the energy spectra of the true distribution with generated samples. Spectrum noise achieves more accurate spectral estimation and maintains this accuracy as resolution increases using a small, fixed amount of integration steps, while white noise performance degrades with grid refinement despite using significantly more integration steps.

\paragraph{Beyond ``white noise vs.\ matched noise'': sweeping the noise smoothness}
The white-noise baseline is an extreme case ($s_0=0$) of our spectrum-mismatch analysis. To check that the message of Section~\ref{sec-The need for numerical efficiency} is not specific to that extreme, we sweep the noise smoothness exponent $s_0\in\{0, 1, 2, 3\}$ at fixed data smoothness $s_1=3$ (so $s_0\le s_1$ throughout, satisfying the rougher-than-data condition). 

\begin{figure}[ht]
    \centering
    \includegraphics[width=0.78\linewidth]{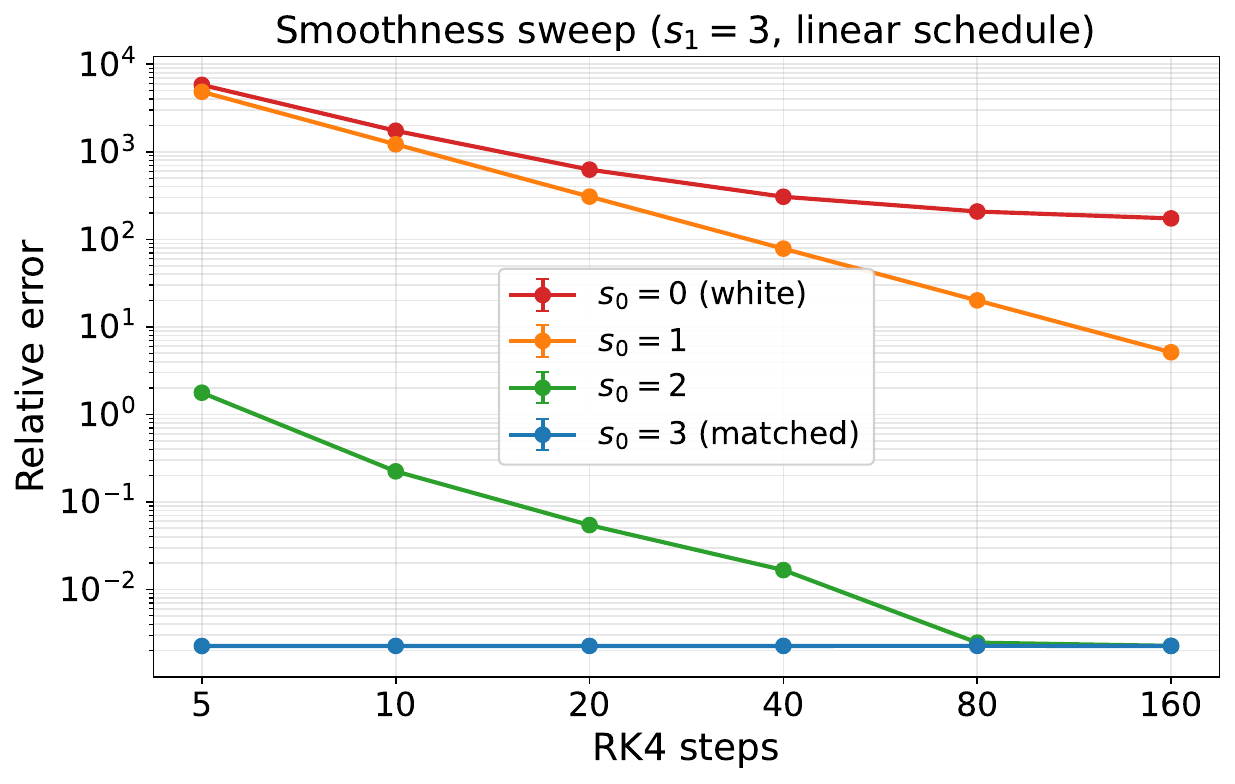}
    \caption{Smoothness sweep: high-band relative error vs.\ RK4 step count, for $s_0\in\{0,1,2,3\}$ at fixed data smoothness $s_1=3$. Horizontal axis: RK4 step count (log). Vertical axis: relative error in $k\ge 24$ (log). The matched case $s_0=s_1=3$ is at the accuracy floor (set by finite-sample error of the empirical spectrum) by $5$ steps. Intermediate $s_0\in\{1,2\}$ interpolate.}
    \label{fig:smoothness-sweep}
\end{figure}

Table~\ref{tab:smoothness-sweep} reports the high-wavenumber relative spectrum error at $128\times 128$ (which is the most significant part of the errors) as a function of integration budget. The error decreases monotonically as $s_0\to s_1$. The matched-spectrum case ($s_0=s_1=3$) is already at the accuracy floor at $5$ RK4 steps; intermediate roughness levels interpolate; the white-noise extreme stays $\sim 200$ even with $80$ steps. Figure~\ref{fig:smoothness-sweep} plots the same data as a function of step count.

\begin{table}[ht]
\centering
\small
\begin{tabular}{ccccc}
\toprule
RK4 steps & $s_0{=}0$ (white) & $s_0{=}1$ & $s_0{=}2$ & $s_0{=}3$ (matched)\\
\midrule
$5$   & $5.8\!\times\!10^{3}$  & $4.9\!\times\!10^{3}$  & $1.77$               & $\mathbf{2.3\!\times\!10^{-3}}$ \\
$10$  & $1.7\!\times\!10^{3}$  & $1.2\!\times\!10^{3}$  & $0.22$               & $\mathbf{2.3\!\times\!10^{-3}}$ \\
$20$  & $626$                  & $308$                  & $0.054$              & $\mathbf{2.3\!\times\!10^{-3}}$ \\
$40$  & $307$                  & $78.3$                 & $0.017$              & $\mathbf{2.3\!\times\!10^{-3}}$ \\
$80$  & $208$                  & $20.1$                 & $2.5\!\times\!10^{-3}$ & $\mathbf{2.3\!\times\!10^{-3}}$ \\
\bottomrule
\end{tabular}
\caption{Relative energy-spectrum error $\frac{1}{|B|}\sum_{k\in B}|S_{\rm gen}(k)-S_{\rm truth}(k)|/S_{\rm truth}(k)$ over the high-wavenumber band $B=\{k\geq 24\}$ at $128\times 128$, varying noise smoothness $s_0$ at fixed data smoothness $s_1=3$. Both noise and data are Mat\'ern Gaussians with $\tau=1$; data $\sigma_1^2=(4\pi^2+\tau_1^2)^{s_1}$; noise $\sigma_0=1$ for $s_0=0$ and $\sigma_0^2=(4\pi^2+\tau_0^2)^{s_0}$ otherwise. Mean over five seeds with $500$ samples each; relative standard deviations are below $1\%$ of the mean throughout (omitted for compactness).}
\label{tab:smoothness-sweep}
\end{table}

\subsubsection{Example: Invariant distributions of stochastic Allen--Cahn}
\label{sec-example-stochastic Allen--Cahn}
We then consider an infinite-dimensional probability measure over continuous functions on the unit interval $[0,1]$, with density formally given by
\begin{equation}
\exp\left(-\int_0^1 \left(\frac{1}{2}(\partial_x u(x))^2 + V(u(x))\right){\rm d}x\right)\, ,
\end{equation}
where $V(u) = (1 - u^2)^2$ is a double-well potential. This represents the invariant distribution of the stochastic Allen--Cahn equation
\begin{equation}
\partial_t u = \partial_{xx}u - V'(u) + \sqrt{2}\eta\, ,
\end{equation}
subject to natural boundary conditions and driven by space-time white noise $\eta$. 

The resulting distribution is bimodal, with sample realizations typically exhibiting approximately constant profiles near $u = \pm 1$. We discretize the spatial domain using finite differences on $N$ equispaced grid points, yielding an $N$-dimensional probability distribution. Samples $x_1$ from this distribution are generated using ensemble MCMC algorithms \cite{chen2025new}, with $5\times 10^4$ training samples; the matched-spectrum noise is constructed analytically from the known Gaussian component.

We compare two choices of Gaussian noise: white noise and spectrum noise that matches the Gaussian component $\exp(-\int_0^1 \frac{1}{2}(\partial_x u(x))^2\,{\rm d}x)$ for mean zero functions. Using the interpolant $I_t = \alpha_t z + \beta_t x_1$ with linear schedule $\alpha_t = 1-t$ and $\beta_t = t$, we construct ODE generative models solved via RK4 schemes.

\begin{figure}[ht]
    \centering
    \includegraphics[width=0.32\linewidth]{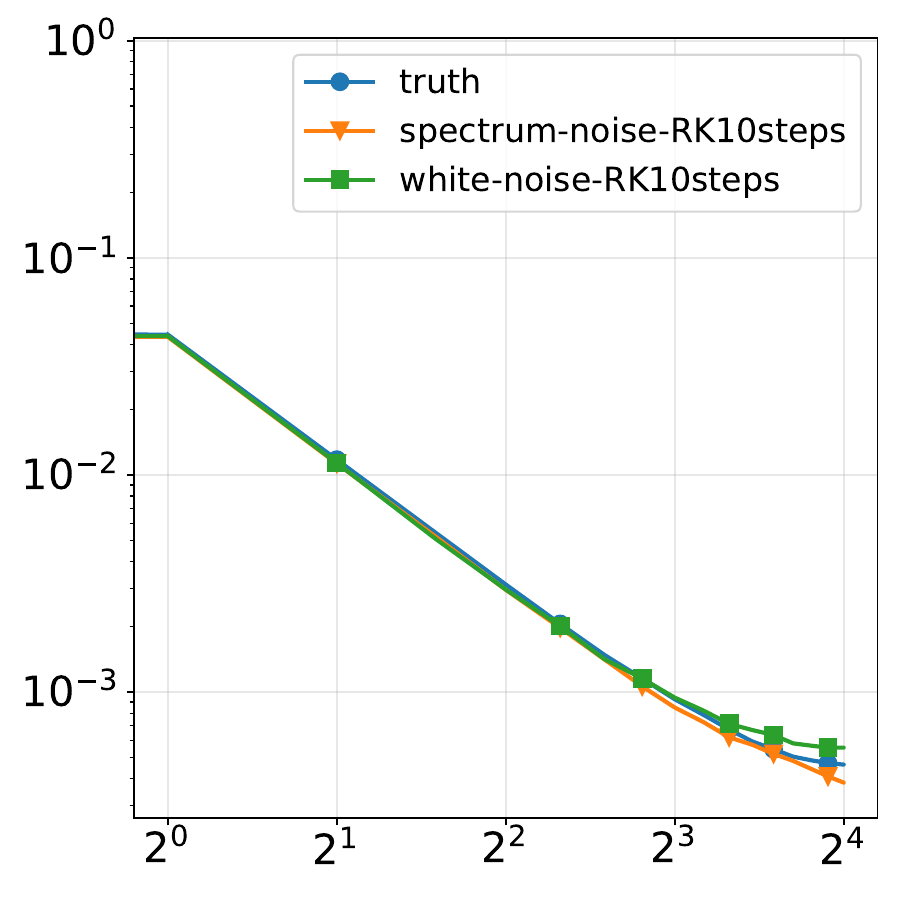}
    \includegraphics[width=0.32\linewidth]{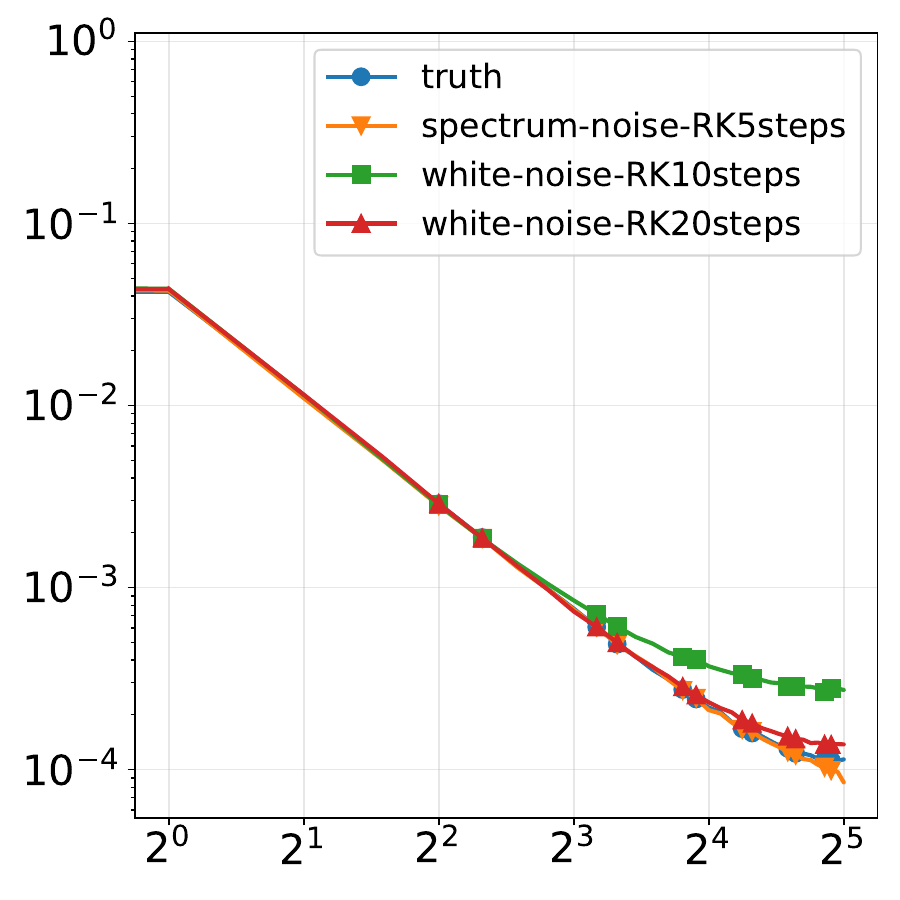}
    \includegraphics[width=0.32\linewidth]{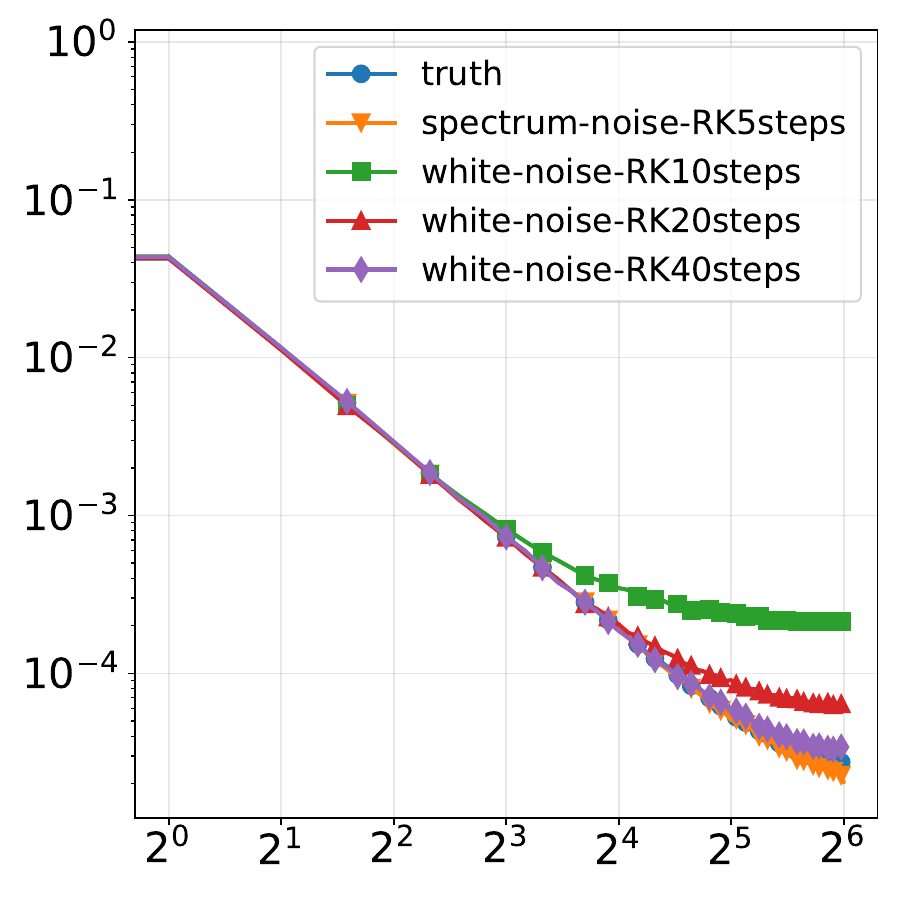}
    \caption{Energy spectra of stochastic Allen--Cahn invariant distributions: comparison between ground truth, spectrum noise generation (5 RK4 steps), and white noise generation (10, 20, or 40 RK4 steps). Left: $N=32$; middle: $N=64$; right: $N=128$.}
    \label{fig:allen-cahn-spectrum}
\end{figure}

Figure \ref{fig:allen-cahn-spectrum} demonstrates that for this mildly non-Gaussian distribution, using spectrum noise matched to the Gaussian component achieves superior accuracy in energy spectra that remains robust across different resolutions, while white noise performance degrades and requires substantially more integration steps.

\subsubsection{Failure example: Invariant distributions of stochastic Navier--Stokes}
\label{sec-failure-example-navier-stokes}

Spectrum noise performs favorably on the Gaussian and Allen--Cahn examples above because the fine-scale structure is well captured by such noise. However, for general distributions, simply using Gaussian noise to mimic the second-order statistics of the data distribution may lead to failures.

As an illustrative example, consider the 2D Navier--Stokes equations with random forcing on the torus $\mathbb{T}^2 = [0,2\pi]^2$. Using the vorticity formulation, the equation can be expressed as:
\begin{equation}
    \label{eq:2D_vorticity_NS}
    \mathrm{d}\omega + v \cdot \nabla\omega\,\mathrm{d}t = \nu \Delta\omega\,\mathrm{d}t - \alpha\omega\,\mathrm{d}t + \varepsilon\,\mathrm{d}\eta\, .
\end{equation}
Here, $v = \nabla^{\perp} \psi = (-\partial_y\psi, \partial_x\psi)$ represents the velocity field derived from the stream function $\psi$, which satisfies $-\Delta \psi = \omega$. The term $\mathrm{d}\eta$ denotes white-in-time random forcing acting on a finite set of Fourier modes, while $\nu$, $\alpha$, and $\varepsilon > 0$ are physical parameters; we take $\nu=10^{-3}$, $\alpha=0.1$, $\varepsilon = 1$, and other parameters following \cite{chen2024probabilistic}. For this choice, equation~\eqref{eq:2D_vorticity_NS} is rigorously proven to be ergodic with a unique invariant measure~\cite{hairer2006ergodicity}. 
\begin{figure}[ht]
    \centering
    \includegraphics[width=0.32\linewidth]{images/NSsample1.pdf}
    \includegraphics[width=0.32\linewidth]{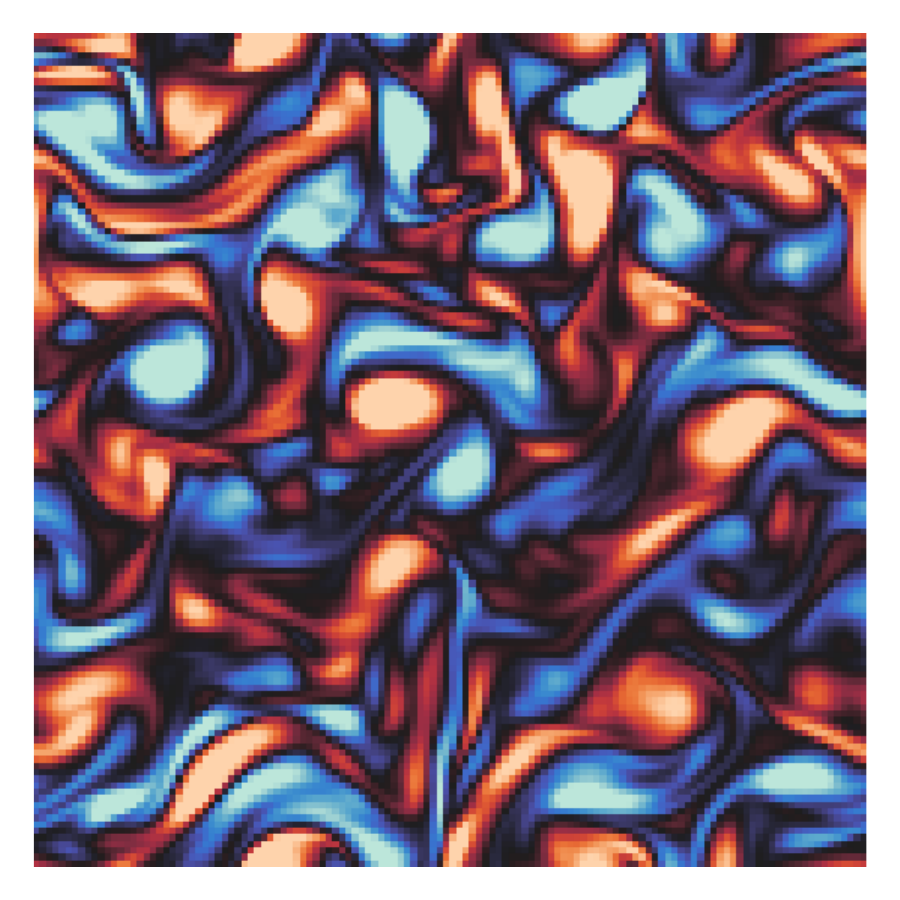}
    \includegraphics[width=0.32\linewidth]{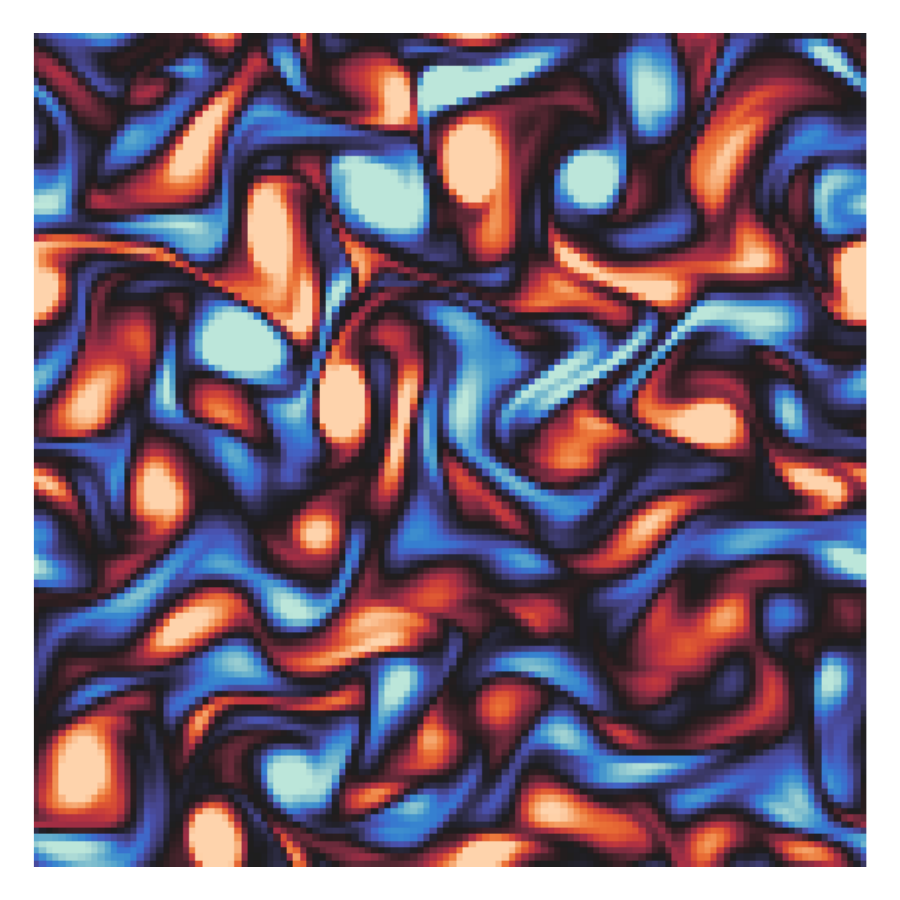}
     \includegraphics[width=0.32\linewidth]{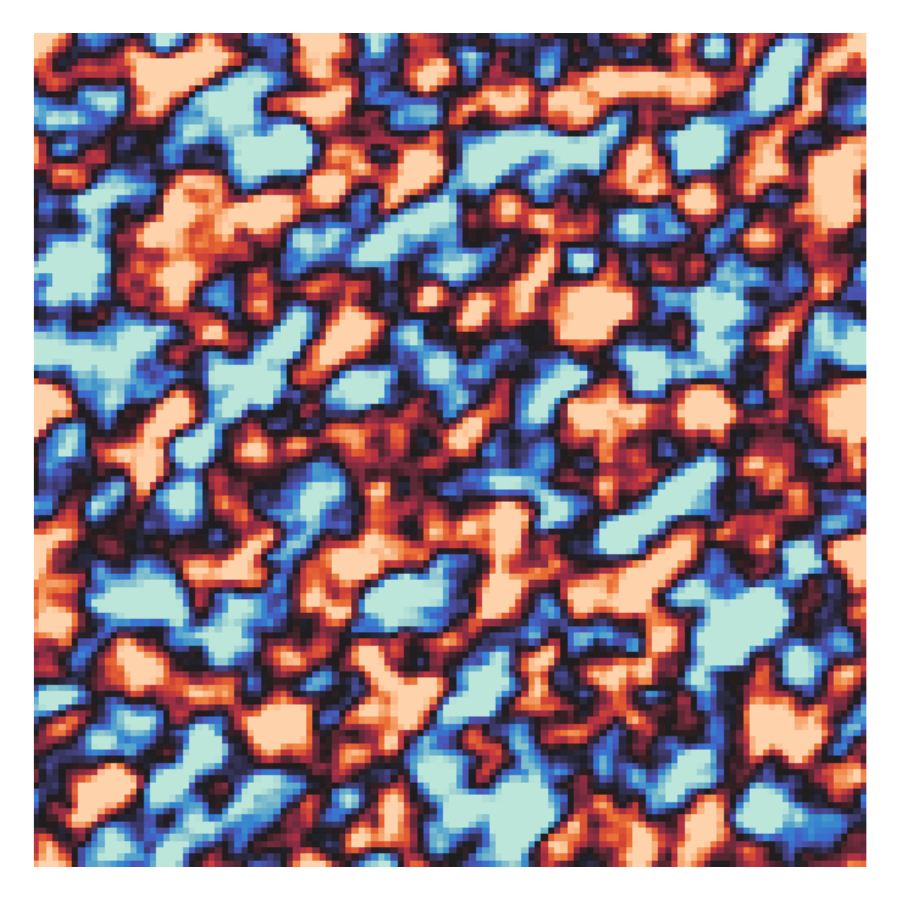}
    \includegraphics[width=0.32\linewidth]{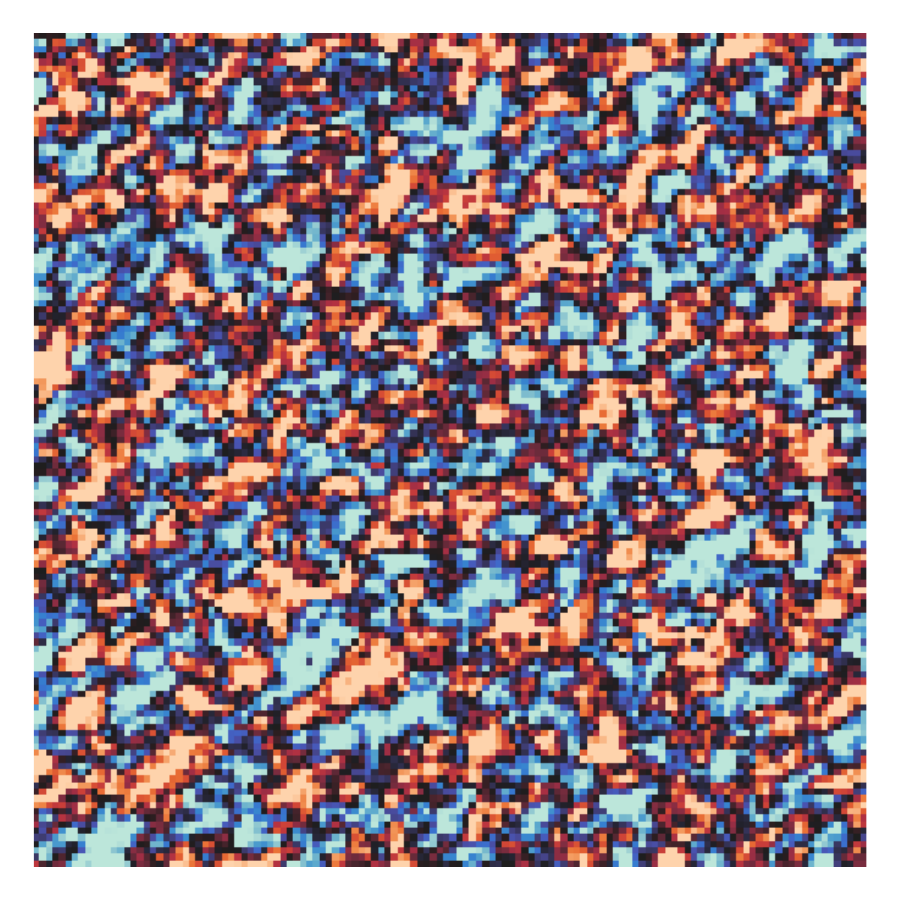}
    \includegraphics[width=0.32\linewidth]{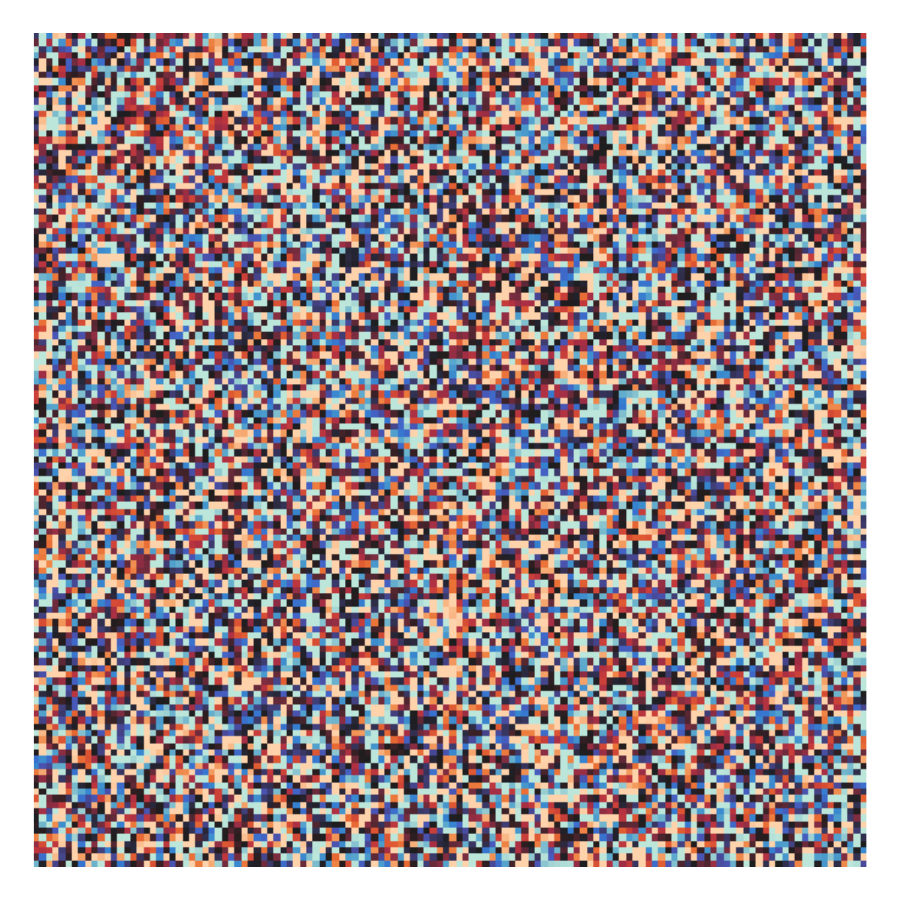}
    \caption{\textbf{Up}: Three samples drawn from the invariant distribution of the stochastically forced Navier--Stokes. \textbf{Down}: three types of noises used for constructing generative models: Gaussian with the same spectrum behavior as the invariant distribution, Gaussian with a rougher spectrum (multiply the Fourier coefficient by $k=\|m\|_2$ at wavenumber $m$), and white noise. All are at the resolution $128\times 128$.}
    \label{fig:samples-nse-and-noises}
\end{figure}

We generate the Navier--Stokes data by long-time simulation on a fine grid, yielding $\sim 9\times 10^4$ training vorticity snapshots ($\sim 10^5$ in total with a $90$/$10$ training/test split); the per-mode variance defining the spectrum noise is estimated empirically from $5\times 10^3$ of these snapshots. Figure \ref{fig:samples-nse-and-noises} shows samples from the invariant distribution along with three noise types that we will use for constructing the ODE generative models. The spectrum noise matches the estimated Fourier spectrum of the data by using empirically determined variances for each Fourier coefficient. The rougher spectrum noise multiplies each Fourier coefficient by the wavenumber magnitude $k=\|m\|_2$ to create a rougher spectral profile. Finally, we include standard
white noise for comparison.

\begin{figure}[ht]
    \centering
    \includegraphics[width=0.32\linewidth]{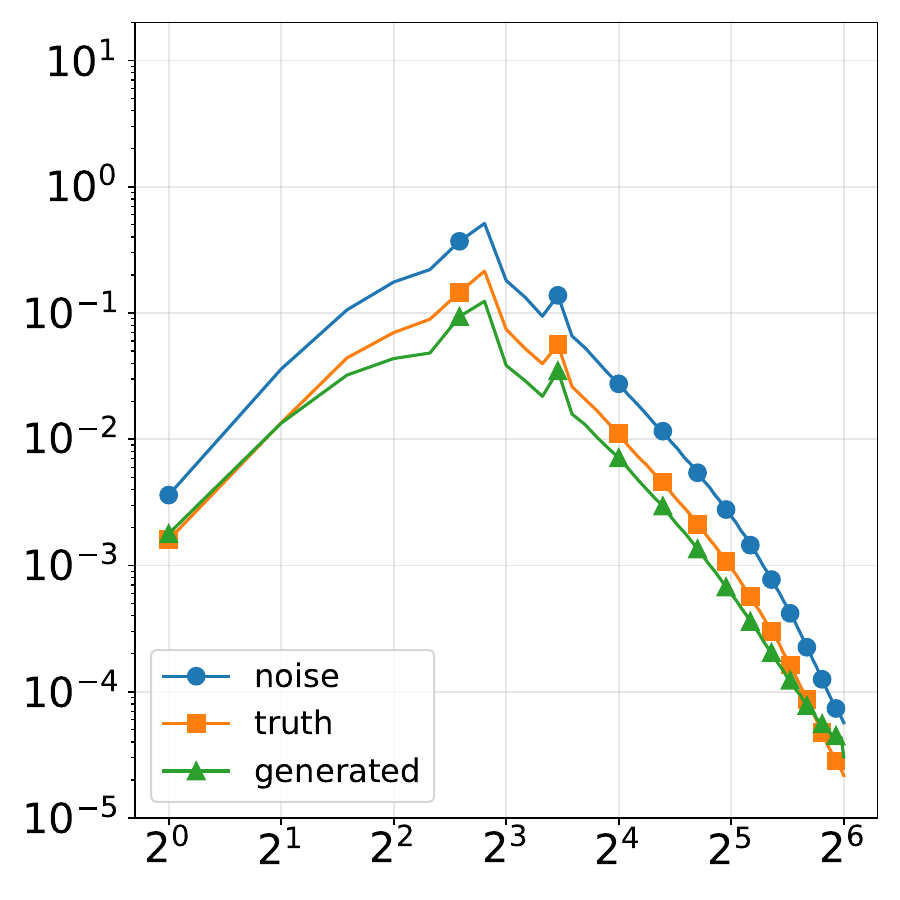}
    \includegraphics[width=0.32\linewidth]{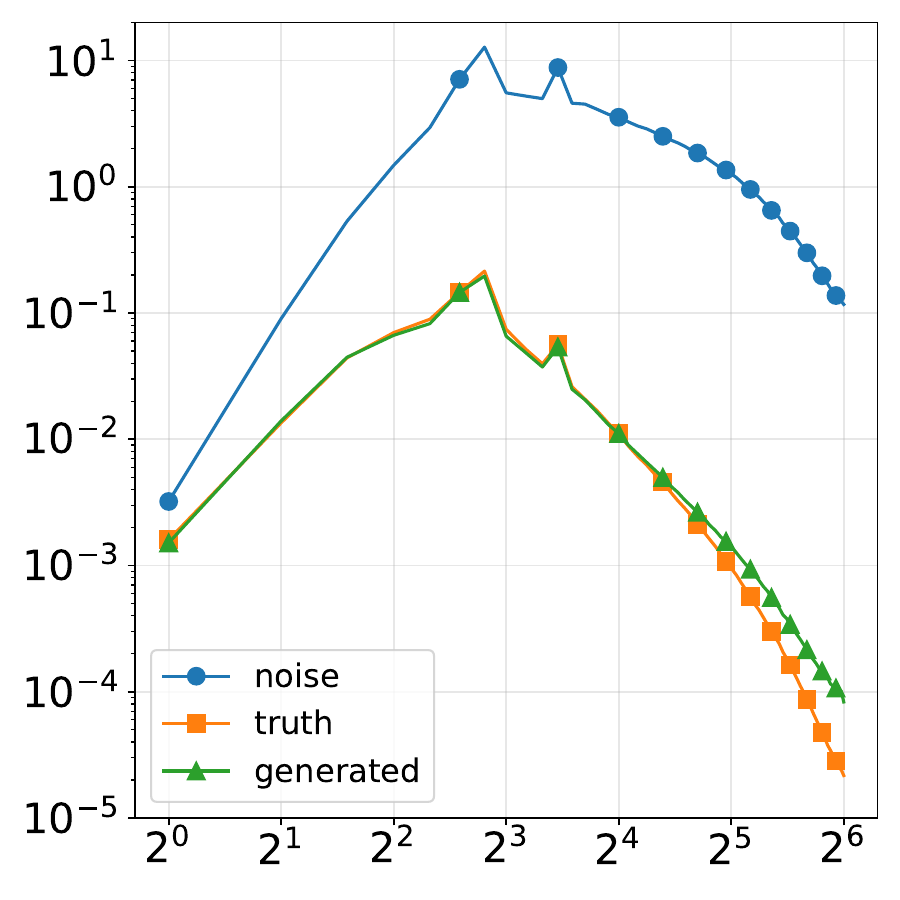}
    \includegraphics[width=0.32\linewidth]{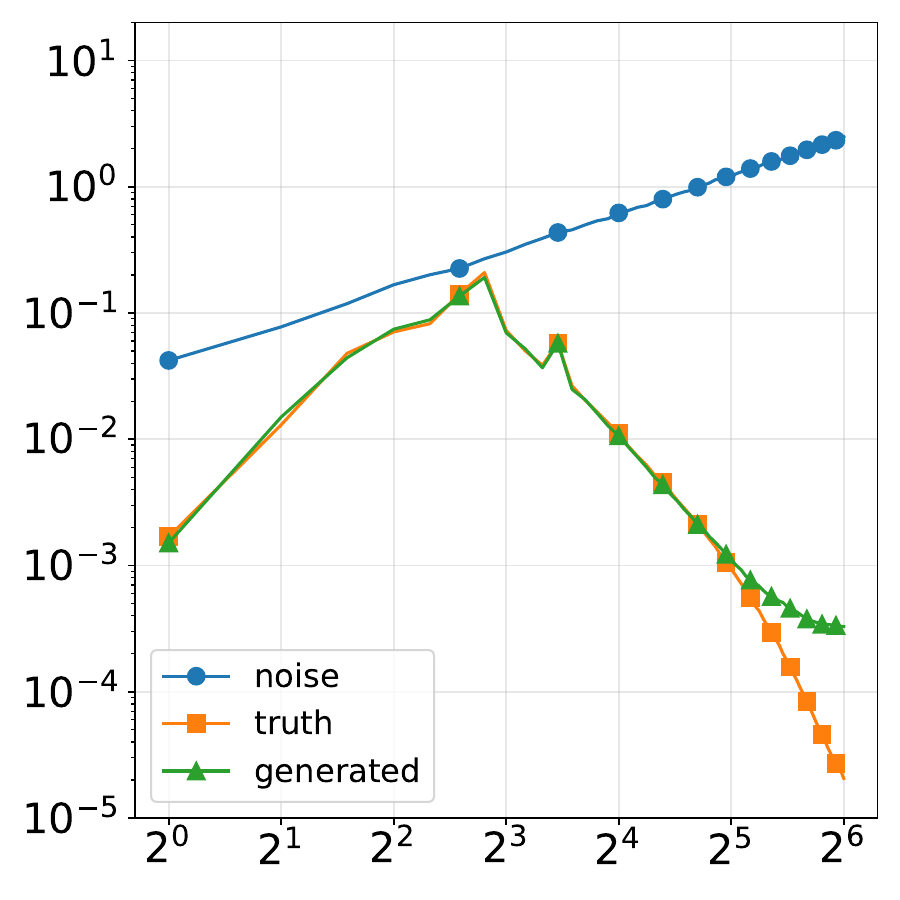}
    \caption{Enstrophy spectrum at resolution $128\times 128$ for the three noise choices in Figure~\ref{fig:samples-nse-and-noises}. \textbf{Left:} matched-spectrum noise. \textbf{Middle:} rougher spectrum noise obtained by multiplying the matched per-mode standard deviation by the wavenumber magnitude $k=\|m\|_2$. \textbf{Right:} white noise. Each panel plots three curves on the same axes: the truth enstrophy spectrum, the noise spectrum, and the spectrum of $500$ generated samples, all averaged over five random seeds. All cases use $10$ RK4 integration steps with the linear schedule. The horizontal axis is the wavenumber magnitude $k$ on a log scale; the vertical axis is enstrophy $S(k)$ on a log scale.}
    \label{fig:generated-results-spectrum}
\end{figure}

 Using the interpolant $I_t = \alpha_t z + \beta_t x_1$ with the three noise types and the linear schedule $\alpha_t = 1-t$, $\beta_t = t$, we train an ODE generative model for each and integrate by RK4. The enstrophy spectra of the generated samples are reported in Figure~\ref{fig:generated-results-spectrum}. 
 
The matched-spectrum noise performs much worse than the rougher noise, and adding more RK4 steps does not help (with the UNet architecture held fixed). Two factors may combine to make this configuration difficult. First, the Navier--Stokes invariant measure is strongly non-Gaussian, so matching only its second-order statistics misses higher-order structure such as intermittency and inter-mode correlations. Second, the bounded-$V$-support sufficient condition of Proposition~\ref{prop-lip-CM-space} is not available for the matched-spectrum noise (the data is not in its Cameron--Martin space; we verify this empirically in Appendix~\ref{appendix-validate-prop33}.A), so this proposition cannot be used to support the Navier--Stokes setting, and the drift $b_t$ appears harder to learn at this $128\times 128$ resolution. The experiments thus indicate that, for challenging problems where precise knowledge of the data's fine-scale structure is not available, rougher-than-data noise is typically needed.

\section{Numerical Efficiency through Design of Interpolation Schedules}
\label{sec-Numerical Efficiency through Design of Interpolation Schedules}
The examples in the previous section show that we may have to use rougher noise in practice, such as in the stochastic Navier--Stokes experiment. As discussed in Section \ref{sec-The need for numerical efficiency}, using rougher noise leads to numerical ill-conditioning when solving the ODE as the grid size increases. In fact, we already observe less accurate estimation of the enstrophy spectrum at high Fourier modes in Figure \ref{fig:generated-results-spectrum}. 

In this section, we show that when using rougher noise, it is possible to design a scale-adaptive interpolation schedule that could improve the numerical conditioning and leads to substantially better estimation with the same number of discretization steps.

\subsection{Motivating study in the case of Gaussian measures}
\label{sec-Motivating study in the case of Gaussian measures}
Returning to the Mat\'ern-like Gaussian measure of Proposition~\ref{prop-gaussian-measure} with $C_0 = \sigma_0^2(-\Delta + \tau_0^2 \mathrm{I})^{-s_0}$ and $C_1 = \sigma_1^2(-\Delta + \tau_1^2 \mathrm{I})^{-s_1}$, we first demonstrate the difficulty posed by the standard linear schedule $\alpha_t = 1 - t$, $\beta_t = t$. The operator
\[B(t) = ((t-1) + t C_1C_0^{-1}) ((t-1)^2 + t^2C_1C_0^{-1})^{-1}\]
has a diagonal representation in Fourier space. For mode $m \in \mathbb{Z}^2_+ \backslash \{0\}$, we have 
\[\tilde{B}(t;m) = \frac{(t-1)+t \mu_m}{(t-1)^2 + t^2 \mu_m}\, , \]
where
\[\mu_m = \frac{\sigma_1^2 (4\pi^2\|m\|^2_2 + \tau_1^2)^{-s_1} }{\sigma_0^2 (4\pi^2\|m\|^2_2 + \tau_0^2)^{-s_0}} \, .\]

Consider the white noise case with $s_0 = 0$ and $\sigma_0 = \sigma_1$, yielding $\mu_m = (4\pi^2\|m\|^2_2 + \tau_1^2)^{-s_1}$. Since $\lim_{m\to \infty} \tilde{B}(t;m) = -\frac{1}{1-t}$, we require smaller time stepsizes as $t$ approaches 1.
While early stopping at some $t < 1$ is possible in principle, high-frequency modes with smaller magnitudes $(4\pi^2\|m\|^2_2 + \tau_1^2)^{-s_1}$ in the data require integration closer to $t = 1$ for fidelity. In Fourier space, the interpolant satisfies $\tilde{I}_t(m) \sim \sfN(0, \sigma_1^2 t^2 (4\pi^2\|m\|^2_2 + \tau_1^2)^{-s_1} + \sigma_1^2(1-t)^2)$. For good relative accuracy in the $m$-th mode, we must integrate until $t$ such that
\[(1-t)^2 \sim (4\pi^2\|m\|^2_2 + \tau_1^2)^{-s_1} \, .\]

At this point, the Lipschitz constant of the drift for the $m$-th mode, approximately $\frac{1}{1-t}$, scales as $(4\pi^2\|m\|^2_2 + \tau_1^2)^{s_1/2}$, growing polynomially with $m$. Consequently, we must decrease the stepsize at rate $(4\pi^2\|m\|^2_2 + \tau_1^2)^{-s_1/2}$, leading to significantly increased computational cost when capturing fine-scale modes.

However, this issue can be addressed using wavenumber-dependent interpolation schedules. In fact, for the Mat\'ern-like example, we seek a wavenumber-dependent linear interpolation in Fourier space:
\[\tilde{I}_t(m) = \alpha_t(m) \tilde{x}_0(m) + \beta_t(m)\tilde{x}_1(m)\, . \]
The drift in Fourier space is diagonal, with per-mode coefficient
\begin{equation*}
    \begin{aligned}
        \tilde{b}_t(m) &= \frac{\dot\alpha_t(m) \alpha_t(m)c_0(m) + \dot\beta_t(m) \beta_t(m) c_1(m)}{\alpha_t^2(m) c_0(m) + \beta_t^2(m) c_1(m)} \\
        & = \frac{1}{2}\frac{\rm d}{{\rm d} t} \log (\alpha_t^2(m) c_0(m) + \beta_t^2(m) c_1(m))\, ,
    \end{aligned}
\end{equation*}
where $c_0(m) = \sigma_0^2(4\pi^2\|m\|^2_2 + \tau_0^2)^{-s_0}$ and $c_1(m) = \sigma_1^2(4\pi^2\|m\|^2_2 + \tau_1^2)^{-s_1}$. 

We can choose $\alpha_t(m), \beta_t(m)$ such that
\[ \log (\alpha_t^2(m) c_0(m) + \beta_t^2(m) c_1(m)) = (1-t)\log c_0(m) + t \log c_1(m)\, . \]
A particular analytic solution is 
\[\alpha_t(m) = \sqrt{\frac{(c_1(m)/c_0(m))^t-c_1(m)/c_0(m)}{1-c_1(m)/c_0(m)}},\ \  \beta_t(m) = \sqrt{\frac{1-(c_1(m)/c_0(m))^t}{1-c_1(m)/c_0(m)}} \, . \]

For this choice, we obtain $\tilde{b}_t(m) = \frac{1}{2}\log \frac{c_1(m)}{c_0(m)}$, which, in the above Mat\'ern-like example, depends on $\|m\|_2$ only logarithmically. Thus, the Lipschitz constant of the drift increases only logarithmically with respect to $\|m\|_2$, yielding an exponential improvement compared to the linear schedule $\alpha_t = 1-t, \beta_t = t$.

While the above discussion requires a wavenumber-dependent (non-scalar) schedule that may be difficult to implement in general, we demonstrate below that a scalar schedule can achieve a similar exponential improvement in the Gaussian case. We write the smallest eigenvalue below as $\lambda^\star$.
\begin{proposition}
\label{prop-gaussian-design-schedule}
Consider the interpolant process
\[I_t = \alpha_t z + \beta_t x_1\]
where $z \sim \sfN(0,C_0)$ and $x_1 \sim \sfN(0,C_1)$ are Gaussian measures in $\bR^d$. We assume $C_0$ and $C_1$ are simultaneously diagonalizable. Let the eigenvalues of $C_1C_0^{-1}$ be $1\geq \lambda_1 \geq  ... \geq \lambda_d>0$. Let $\lambda^\star := \lambda_d$. Then, taking the scalar interpolation schedule
\begin{equation}
\label{eqn-high-D-gaussian-alpha-beta}
    \alpha_t =\sqrt{\frac{\lambda^\star-(\lambda^\star)^t}{\lambda^\star-1}}, \ \ \beta_t =   \sqrt{\frac{(\lambda^\star)^t - 1}{\lambda^\star-1}}\, ,
\end{equation}
we have $\|\nabla b_t(x)\|_2 = \frac{1}{2}|\log \lambda^\star|$ for every $x\in\bR^d$ and every $t\in(0,1)$.
\end{proposition}
The proof of this proposition is in Appendix \ref{sec-Proof for Gaussian Target Measures}. We note the above schedule is not in $C^1([0,1])$ as it is not smooth near the boundary, but in practice we often set a buffer near the boundary which effectively smooths the schedule out. Appendix~\ref{appendix-validate-prop33}.B numerically verifies the predicted Lipschitz constant $\frac12|\log\lambda^\star|$ alongside the linear-schedule behavior of Proposition~\ref{prop-gaussian-measure}. Let us make some remarks. Consider the Mat\'ern-like example restricted to the first $d$ modes, with the per-mode ratio
\[\lambda_m \;=\; \frac{\sigma_1^2 (4\pi^2\|m\|^2_2 + \tau_1^2)^{-s_1} }{\sigma_0^2 (4\pi^2\|m\|^2_2 + \tau_0^2)^{-s_0}}.\]
Consider the case $s_0 = 0, s_1 > 0, \sigma_0=\sigma_1$. We have $\lambda^\star = \lambda_m$ for the $m$ that achieves the largest $\|m\|_2$ among the first $d$ modes. The Lipschitz constant of the drift depends only on $|\log \lambda^\star|$, which scales logarithmically with $\|m\|_2$ rather than polynomially as in the linear schedule case. This logarithmic dependence means that fine-scale modes can be captured without significantly more computational effort in time integration. A more refined wavenumber-dependent variant (using $\lambda_m$ per mode instead of the global $\lambda^\star$) is in principle per-mode optimal; numerically the simpler scalar schedule loses at most a small constant factor in step count, and we defer the comparison to Appendix~\ref{appendix-validate-prop33}.C.

\subsection{Improving the result for stochastic Navier--Stokes}
\label{sec-Improving-NS}
We use the insights derived from the Gaussian setting to improve the generative modeling of the stochastically forced Navier--Stokes example in Section \ref{sec-failure-example-navier-stokes}.

We choose white noise in the construction of our generative models. The schedule parameter $\lambda^\star$ in \eqref{eqn-high-D-gaussian-alpha-beta} is the smallest eigenvalue of $C_1 C_0^{-1}$ in the Gaussian theory. For a non-Gaussian target it is not directly accessible, so we use a data-driven proxy: at the finest resolved wavenumber $k_{\max}$,
\begin{equation}
\label{eq:auto-lambda-ns}
\lambda^\star = \frac{S_{\rm data}(k_{\max})}{S_{\rm noise}(k_{\max})},
\end{equation}
where $S_{\rm data}, S_{\rm noise}$ are the radially-averaged enstrophy spectra of the data and the noise, respectively. This rule encodes the same intuition as the eigenvalue ratio in the Gaussian case and is computed once from the training set. Empirically it gives $\lambda^\star \approx 3\times 10^{-4}$ at $64\times 64$ and $\lambda^\star \approx 10^{-5}$ at $128\times 128$. The designed schedule is applied at inference time via the standard transfer formula relating drifts of two scalar schedules \cite{kingma2021variational, karras2022elucidating, chen2025lipschitz}, so both schedules use the same trained drift; the ODE is integrated with a fixed-step RK4 method on a uniform grid in $[t_{\min}, t_{\max}] = [10^{-3}, 1-10^{-3}]$.

\begin{figure}[ht]
    \centering
    \includegraphics[width=0.32\linewidth]{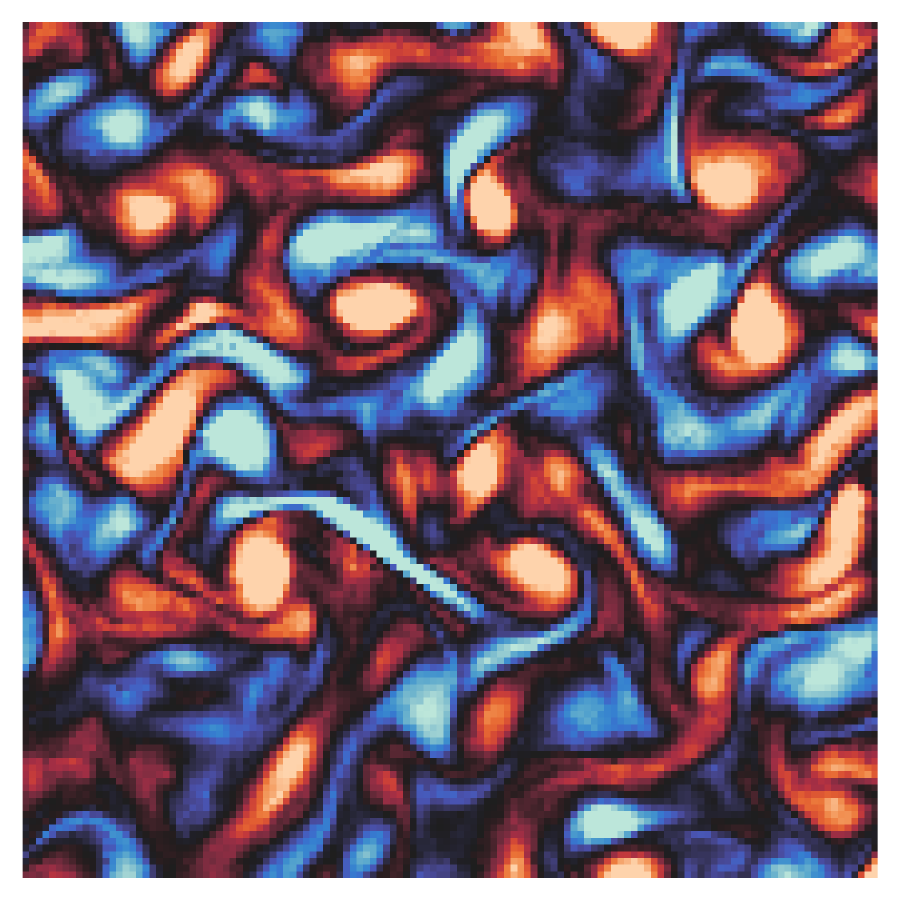}
    \includegraphics[width=0.32\linewidth]{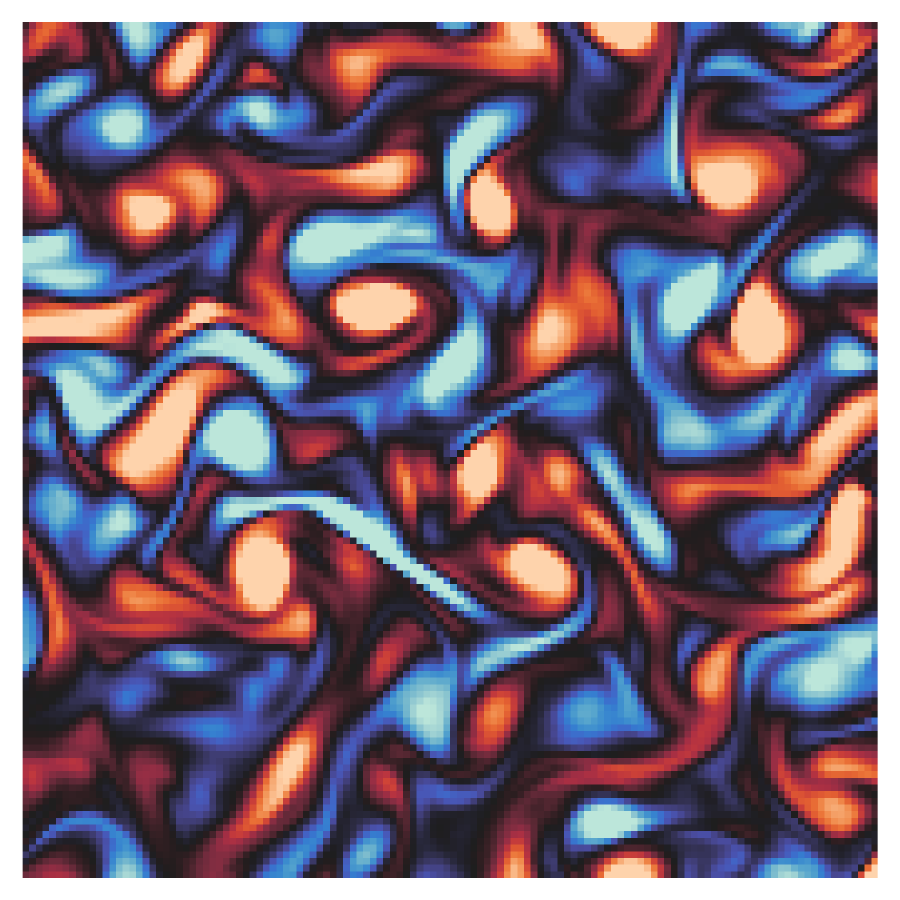}
    \includegraphics[width=0.32\linewidth]{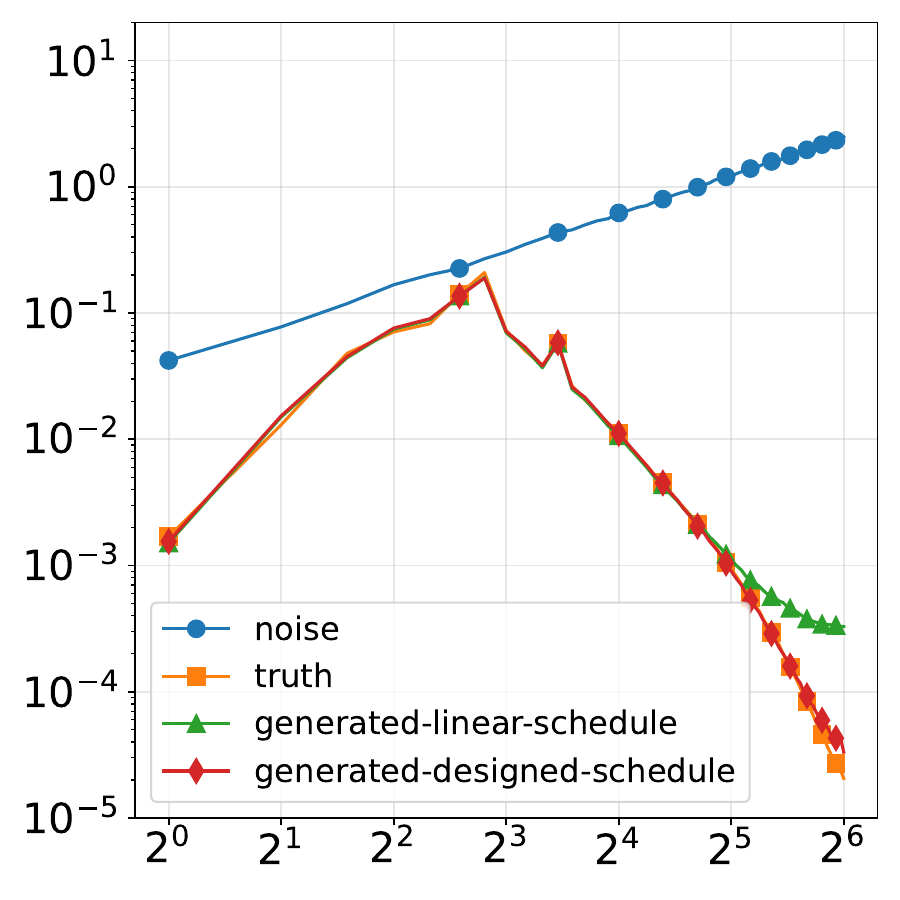}
    \caption{Experiments on stochastically forced Navier--Stokes using white noise in the generative model. \textbf{Left:} sample generated with the linear schedule. \textbf{Middle:} sample generated with the designed schedule \eqref{eqn-high-D-gaussian-alpha-beta}, using the data-driven $\lambda^\star\approx 10^{-5}$. \textbf{Right:} ensemble-averaged enstrophy spectra of $500$ generated samples per schedule against the truth and the noise. In all cases we use $10$ RK4 integration steps; resolution $128\times 128$. The designed schedule yields visibly smoother samples without spurious fine-scale artefacts and reproduces the enstrophy spectrum across all frequencies, while the linear schedule overestimates fine-scale enstrophy by orders of magnitude.}
    \label{fig:NSE-generated-results-spectrum-compare-designed-schedule}
\end{figure}

Figure~\ref{fig:NSE-generated-results-spectrum-compare-designed-schedule} demonstrates that with only $10$ RK4 steps, the designed schedule produces qualitatively cleaner samples and an enstrophy spectrum that tracks the truth across all wavenumbers, in sharp contrast to the linear schedule and to the matched-spectrum noise (see Section~\ref{sec-failure-example-navier-stokes}). The visual improvement is further supported by two quantitative diagnostics that we report below: (i) spectrum errors split by low/mid/high wavenumber bands, and (ii) non-Gaussian statistics (flatness and gradient kurtosis) that probe intermittent fine-scale structure.

\paragraph{Spectrum error by wavenumber band}
To probe multiscale fidelity, we report the spectrum error split into low/mid/high wavenumber bands rather than a single global average. For a band $B\subset \{1,\dots,k_{\max}\}$ we compute the unweighted mean relative error $\frac{1}{|B|}\sum_{k\in B}|S_{\rm gen}(k)-S_{\rm truth}(k)|/|S_{\rm truth}(k)|$: low ($k<8$), mid ($8\le k<24$), and high ($k\ge 24$). Table~\ref{tab:ns-perband} reports mid- and high-band errors at $64\times 64$ and $128\times 128$ as the integrator budget is varied (low-band errors are similar across schedules). The improvement is most dramatic in the high band, which is precisely where the linear-schedule analysis of Section~\ref{sec-The need for numerical efficiency} predicts the largest stiffness: at $128\times 128$ and $10$ RK4 steps, the linear schedule's high-band error is $\sim 355\%$ while the designed schedule reaches $\sim 16\%$, a $\sim 20\times$ reduction. Even with $5\times$ as many RK4 steps ($50$ vs $10$, a $5\times$ NFE/wall-clock penalty), the linear schedule's high-band error ($\sim 26\%$) still exceeds the designed schedule's at $10$ steps. A few entries are mildly non-monotone in step count due to the statistical training error of the learned drift and network representation.

\begin{table}[ht]
\centering
\small
\begin{tabular}{llcc}
\toprule
Resolution & Method (steps) & Mid ($8\le k<24$) & High ($k\ge 24$) \\
\midrule
$64\times 64$ & Linear (10)   & $0.060 \pm 0.001$ & $0.421 \pm 0.011$ \\
$64\times 64$ & Linear (20)   & $0.056 \pm 0.006$ & $0.054 \pm 0.007$ \\
$64\times 64$ & Linear (50)   & $0.066 \pm 0.006$ & $0.032 \pm 0.012$ \\
$64\times 64$ & Designed (10) & $\mathbf{0.033 \pm 0.006}$ & $\mathbf{0.041 \pm 0.004}$ \\
$64\times 64$ & Designed (20) & $\mathbf{0.036 \pm 0.006}$ & $\mathbf{0.014 \pm 0.006}$ \\
$64\times 64$ & Designed (50) & $\mathbf{0.033 \pm 0.006}$ & $\mathbf{0.015 \pm 0.002}$ \\
\midrule
$128\times 128$ & Linear (10)   & $0.039 \pm 0.010$ & $3.549 \pm 0.049$ \\
$128\times 128$ & Linear (20)   & $0.049 \pm 0.011$ & $0.803 \pm 0.013$ \\
$128\times 128$ & Linear (50)   & $0.051 \pm 0.011$ & $0.255 \pm 0.004$ \\
$128\times 128$ & Designed (10) & $\mathbf{0.019 \pm 0.003}$ & $\mathbf{0.163 \pm 0.009}$ \\
$128\times 128$ & Designed (20) & $\mathbf{0.017 \pm 0.006}$ & $\mathbf{0.160 \pm 0.005}$ \\
$128\times 128$ & Designed (50) & $\mathbf{0.017 \pm 0.006}$ & $\mathbf{0.179 \pm 0.005}$ \\
\bottomrule
\end{tabular}
\caption{Relative spectrum errors by wavenumber band at two resolutions ($64\times 64$ and $128\times 128$) and three integrator budgets ($10$, $20$, $50$ RK4 steps). Both methods use a white noise; the designed schedule is applied at inference via the transfer formula. Errors are mean $\pm$ s.d.\ over five independent seeds, $500$ samples each. Bold marks the smaller mean per (resolution, step count, band). The largest gain is in the high band, where the linear-schedule analysis of Section~\ref{sec-The need for numerical efficiency} predicts the largest stiffness.}
\label{tab:ns-perband}
\end{table}

\paragraph{Beyond second-order statistics: intermittency}
Spectra capture only the second-order statistics of the field. Since the invariant measure of \eqref{eq:2D_vorticity_NS} is strongly non-Gaussian, we additionally probe higher-order structure. We report (i) the flatness $F(r) = S_4(r)/S_2(r)^2$ of vorticity increments at scales $r\in\{1,2\}$ pixels, where $S_p(r) = \tfrac12\,\mathbb{E}[|\omega(\cdot+re_1)-\omega(\cdot)|^p] + \tfrac12\,\mathbb{E}[|\omega(\cdot+re_2)-\omega(\cdot)|^p]$ is the order-$p$ increment moment at lag $r$ averaged over the two grid directions $e_1,e_2$ (Appendix~\ref{appendix-reproducibility}), with $F(r)=3$ for a Gaussian field; and (ii) the gradient kurtosis, equal to $F(1)$. The flatness is a standard intermittency diagnostic in turbulence. Recovering it correctly is therefore a stringent test of the generated fine-scale structure, going beyond what the spectrum alone can capture.

Table~\ref{tab:ns-nongaussian} reports these statistics on $128\times 128$ vorticity. The truth ensemble has flatness $\approx 4.95$ at $r=1$ and $\approx 4.34$ at $r=2$, well above the Gaussian baseline of $3$. With $10$ RK4 steps, the designed schedule recovers the $r=1$ flatness within $\sim 3\%$, whereas the linear schedule underestimates it by $\sim 13\%$, narrowing to $\sim 5\%$ at $50$ steps but still wider than the designed schedule's $\sim 3\%$ at $10$ steps. The same pattern holds for the larger increment $r=2$ and for the gradient kurtosis. Thus, although our schedule is derived from purely Gaussian Lipschitz analysis (Section~\ref{sec-Motivating study in the case of Gaussian measures}), the resulting integration scheme reproduces not only second-order but also intermittent fine-scale information of the turbulent flow more faithfully than the linear-schedule baseline.

\begin{table}[ht]
\centering
\small
\begin{tabular}{lccc}
\toprule
Method (steps) & Flatness $r{=}1$ & Flatness $r{=}2$ & Gradient kurtosis \\
\midrule
Truth          & $4.95$            & $4.34$            & $4.95$            \\
\midrule
Linear (10)    & $4.29 \pm 0.02$ & $4.13 \pm 0.02$ & $4.29 \pm 0.02$ \\
Linear (20)    & $4.67 \pm 0.03$ & $4.27 \pm 0.03$ & $4.67 \pm 0.03$ \\
Linear (50)    & $4.69 \pm 0.03$ & $4.28 \pm 0.03$ & $4.69 \pm 0.03$ \\
Designed (10)  & $\mathbf{4.82 \pm 0.03}$ & $\mathbf{4.31 \pm 0.03}$ & $\mathbf{4.82 \pm 0.03}$ \\
Designed (20)  & $\mathbf{4.83 \pm 0.03}$ & $\mathbf{4.32 \pm 0.03}$ & $\mathbf{4.83 \pm 0.03}$ \\
Designed (50)  & $\mathbf{4.83 \pm 0.03}$ & $\mathbf{4.32 \pm 0.03}$ & $\mathbf{4.83 \pm 0.03}$ \\
\bottomrule
\end{tabular}
\caption{Higher-order diagnostics on generated $128\times 128$ vorticity fields. Flatness $F(r) = S_4(r)/S_2(r)^2$ at vorticity-increment scale $r$ measures small-scale intermittency (Gaussian baseline: $F\equiv 3$); the gradient kurtosis is $F(1)$. Means and standard deviations are over five independent seeds with $500$ generated samples per seed; the ``Truth'' row is evaluated on the held-out test set. The designed schedule recovers the truth flatness within a few percent already at $10$ RK4 steps.}
\label{tab:ns-nongaussian}
\end{table}

\paragraph{When to match noise vs.\ use rougher noise}
Across our three test cases, matched-spectrum noise works on the Gaussian and Allen--Cahn targets and fails on Navier--Stokes. A simple, computationally cheap predictor of which regime applies is the fine-scale increment flatness of the data: $F(r) = S_4(r)/S_2(r)^2$ as in the previous non-Gaussian statistics analysis. $F(r)=3$ for a Gaussian field; values $> 3$ indicate fine-scale non-Gaussian behaviors. A matched-spectrum (Gaussian) noise reproduces only the second-order (spectral) statistics, so for strongly intermittent data ($F\gg 3$) it captures the fine-scale energy but not the higher-order structure, which can make the drift harder to learn and integrate. We measure $F(r=1)$ directly on the three test datasets at $128\times 128$ (or $128$ for Allen--Cahn): the Gaussian Mat\'ern field gives $F(r=1) = 3.00$, Allen--Cahn $F(r=1) \approx 2.96$, and Navier--Stokes vorticity $F(r=1) \approx 4.95$ (Table~\ref{tab:ns-nongaussian}). This diagnostic is consistent with the behavior observed across all three: $F(r=1)$ near $3$ for the matched-noise-friendly settings, well above $3$ for the setting where matched noise fails. As a practical guideline, we recommend computing $F(r=1)$ on a small batch of training data: if it is close to $3$, use matched-spectrum + linear schedule; if it is well above $3$, use rougher-than-data noise with the scale-adaptive schedule.

\section{Conclusions}

This work develops principles for designing noise distributions and interpolation schedules in flow-based generative models targeting numerically ill-conditioned distributions with multiscale Fourier spectra. Working in the function-space stochastic-interpolant framework, we study the Lipschitz regularity of the drift to establish well-posedness of the sampling ODE in function space and analyze the step count required by an explicit integrator at any finite resolution. The rougher-than-data condition on the noise is needed for the drift field to be bounded and Lipschitz near $t=0$; absent further design, however, the same condition introduces a terminal-time stiffness that drives the integrator cost up.

We address these competing requirements through two complementary strategies tailored to the available prior knowledge on the data. For distributions whose fine-scale structure is analytically tractable---such as Gaussian random fields and stochastic Allen--Cahn invariant measures that are absolutely continuous with respect to known Gaussian processes---we show that matched noise provides substantial computational advantages while maintaining spectral fidelity. For complex distributions lacking precise fine-scale characterization, such as turbulent Navier--Stokes flows at the given grid scale, we develop scale-adaptive interpolation schedules that enable effective use of rougher noise while preserving numerical stability.

The numerical experiments with both approaches report improvements in efficiency and accuracy across resolutions, and in per-band, flatness, and gradient-kurtosis metrics for Navier--Stokes. We have also articulated a simple data-driven diagnostic for choosing between the two recipes: the fine-scale increment flatness $F(r=1)$ of the data, equal to $3$ for a Gaussian field. Across our three test cases, $F(r=1) \approx 3$ on the matched-noise-friendly settings (Gaussian fields, Allen--Cahn) and $\approx 5$ on Navier--Stokes vorticity, where matched-spectrum noise fails and rougher-than-data noise with the designed schedule is the better choice. Together, these results identify the noise covariance and the time schedule as natural, low-cost design degrees of freedom that can substantially improve flow-based generative models on multiscale scientific data.

Several directions are natural for further improvement. The analysis of scale-adaptive schedule is limited to Gaussian case and extending the design of schedules to more general multiscale choices and conducting systematic analysis is important for the deployment of the strategy in practice. Extension to three-dimensional systems and higher resolutions would demonstrate scalability for modern computational requirements. A more refined analysis of how the training-objective norm interacts with learning efficiency and spectral sensitivity would make the theoretical picture more complete. Integration with physics-informed strategies and exploration of non-Gaussian noise families informed by specific physical processes are also promising avenues for enhancing generative modeling of scientific phenomena.

\appendix
\section{Proofs for Stochastic Interpolants}
\label{appendix:derivation-stochastic-interpolants}

We prove Propositions~\ref{prop-si-bt} (Euclidean) and~\ref{prop-si-bt-functionspace} (function-space). Each proof has three steps: (1) $\mu_t$ satisfies the weak continuity equation; (2) using the assumed unique strong solution of the ODE, $\mathrm{Law}(X_t)$ also satisfies it; (3) by the assumed uniqueness of the weak solution, the two coincide.

\subsection{Proof of Proposition~\ref{prop-si-bt} (Euclidean)}
\label{appendix:proof-Rd}

Throughout this subsection, $\phi\in C^1_c(\mathbb R^d)$ is a smooth compactly supported test function.

\medskip\noindent\textbf{Setup of $b_t$.}~
The random variable $\dot I_t = \dot\alpha_t z + \dot\beta_t x_1$ takes values in $\mathbb R^d$ and is integrable by the finite second-moment assumption on $\mu^*$. The conditional expectation $\mathbb E[\dot I_t\mid I_t=x] = b_t(x)$ is a Borel-measurable function defined $\mu_t$-a.e., and is an element of $L^2(\mathrm d\mu_t;\mathbb R^d)$ since $\mathbb E[\|\dot I_t\|^2]<\infty$.

\medskip\noindent\textbf{Step 1: $\mu_t$ satisfies the weak continuity equation.}~
For fixed $(z,x_1)$, the trajectory $t\mapsto I_t = \alpha_t z + \beta_t x_1$ is $C^1$ in $t$, so the chain rule gives pointwise
$\frac{\mathrm d}{\mathrm dt}\phi(I_t) = \dot I_t\cdot\nabla\phi(I_t)$.
The right-hand side is bounded by $(|\dot\alpha_t|\|z\| + |\dot\beta_t|\|x_1\|)\|\nabla\phi\|_\infty$, integrable in $(t,z,x_1)$. Fubini and the tower property of conditional expectation yield
\begin{equation}
\label{eq:weak-CE-mu-Rd}
\int\phi\,\mathrm d\mu_t \;=\; \int\phi\,\mathrm d\mu_0 \,+\, \int_0^t\!\!\int b_s(x)\cdot\nabla\phi(x)\,\mu_s(\mathrm dx)\,\mathrm ds.
\end{equation}

\medskip\noindent\textbf{Step 2: $\nu_t := \mathrm{Law}(X_t)$ satisfies the same equation.}~
By the hypothesis, the ODE $\dot X_t = b_t(X_t)$ with $X_0\sim\mu_0$ admits a unique strong solution $X_t$. The chain rule gives
$\frac{\mathrm d}{\mathrm dt}\phi(X_t) = b_t(X_t)\cdot\nabla\phi(X_t)$;
integrating and taking expectation,
\begin{equation}
\label{eq:weak-CE-nu-Rd}
\int\phi\,\mathrm d\nu_t \;=\; \int\phi\,\mathrm d\mu_0 \,+\, \int_0^t\!\!\int b_s(x)\cdot\nabla\phi(x)\,\nu_s(\mathrm dx)\,\mathrm ds.
\end{equation}

\medskip\noindent\textbf{Step 3: Uniqueness identifies the marginals.}~
By Steps~1--2, both $\mu_t$ and $\nu_t$ are weak solutions of the continuity equation with the same initial condition $\mu_0$. The hypothesis that $\nu_t$ is the unique such solution then gives $\mu_t = \nu_t$ for every $t\in[0,1]$.

\subsection{Proof of Proposition~\ref{prop-si-bt-functionspace} (function-space)}
\label{appendix:proof-fs}

The structure mirrors the Euclidean proof, with two changes: $\dot I_t,\,I_t$ are now $H$-valued (handled via Bochner integration), and the test functions are cylindrical on $H$.

\medskip\noindent\textbf{Setup of $b_t$.}~
Since $\mathbb E[\|\dot I_t\|_H^2]<\infty$, $\dot I_t$ is Bochner-integrable in $H$. The Bochner conditional expectation $\mathbb E[\dot I_t\mid I_t]$ exists and can be represented by a Borel-measurable $b_t\colon H\to H$ defined $\mu_t$-a.e., with $b_t\in L^2(\mathrm d\mu_t;H)$ \cite{hytonen2016analysis}.

\medskip\noindent\textbf{Cylindrical test functions.}~
A cylindrical test function on $H$ has the form
\[
\phi(x) \;=\; \varphi\big(\langle x,e_1\rangle_H,\ldots,\langle x,e_k\rangle_H\big),\qquad k\in\mathbb N,\;\; e_1,\ldots,e_k\in H,\;\; \varphi\in C^1_c(\mathbb R^k),
\]
with $\nabla\phi(x) = \sum_{i=1}^k (\partial_i\varphi)(\cdots)\,e_i\in H$. For any $C^1$ curve $t\mapsto Y_t$ in $H$, the chain rule reduces to the multivariable chain rule on $\mathbb R^k$ via the inner products $\langle Y_t,e_i\rangle_H$:
\[
\frac{\mathrm d}{\mathrm dt}\phi(Y_t) \;=\; \langle \dot Y_t,\,\nabla\phi(Y_t)\rangle_H.
\]
Cylindrical functions separate Borel probability measures on the separable Hilbert space $H$~\cite{bogachev1998gaussian}.

\medskip\noindent\textbf{Steps 1--3 (function-space).}~
Repeat the Euclidean steps for every cylindrical $\phi$, using the cylindrical chain rule above and the Bochner tower property. Steps~1--2 yield the function-space analogues of~\eqref{eq:weak-CE-mu-Rd}--\eqref{eq:weak-CE-nu-Rd}:
\begin{align}
\label{eq:weak-CE-mu-H}
\int\phi\,\mathrm d\mu_t &\;=\; \int\phi\,\mathrm d\mu_0 \,+\, \int_0^t\!\!\int \langle b_s(x),\nabla\phi(x)\rangle_H\,\mu_s(\mathrm dx)\,\mathrm ds,\\
\label{eq:weak-CE-nu-H}
\int\phi\,\mathrm d\nu_t &\;=\; \int\phi\,\mathrm d\mu_0 \,+\, \int_0^t\!\!\int \langle b_s(x),\nabla\phi(x)\rangle_H\,\nu_s(\mathrm dx)\,\mathrm ds,
\end{align}
where in Step~2 the unique strong solution assumed in the proposition is used. Step~3: by the assumed uniqueness of the weak solution, $\int\phi\,\mathrm d(\mu_t - \nu_t) = 0$ for every cylindrical $\phi$; by separation, $\mu_t = \nu_t$ as Borel measures.

\section{Proof for Gaussian Target Measures}
\label{sec-Proof for Gaussian Target Measures}
\subsection{Proof for Proposition \ref{prop-gaussian-measure}}
\begin{proof}
\medskip\noindent\textbf{Law of $I_t$ and the domain of $B(t)$.}~
By independence and linearity, $I_t = \alpha_t z + \beta_t x_1$ is Gaussian with mean $0$ and covariance $\Sigma_t = \alpha_t^2 C_0 + \beta_t^2 C_1$. The latter is positive (a sum of positive operators), self-adjoint (a sum of self-adjoint operators), and trace-class (a sum of trace-class operators); in particular it is compact. By assumption $C_0$ and $C_1$ are diagonal in a common orthonormal basis $\{e_j\}$, with $C_0 e_j = c_0(j)e_j$ and $C_1 e_j = c_1(j)e_j$, $c_0(j),c_1(j)>0$; hence $\Sigma_t$ is also diagonal, $\Sigma_t e_j = \big(\alpha_t^2 c_0(j) + \beta_t^2 c_1(j)\big)e_j$, with strictly positive eigenvalues for $t\in(0,1)$. In particular $\Sigma_t$ is injective on $H$, so its range is dense, $\overline{\mathrm{Ran}(\Sigma_t)} = \ker(\Sigma_t^*)^\perp = H$, and $\Sigma_t^{-1}\colon\mathrm{Ran}(\Sigma_t)\to H$ is well defined.

\medskip\noindent\textbf{Closed form of $B(t)$.}~
The pair $(I_t, \dot I_t)$ is jointly Gaussian on $H\times H$ with $\mathrm{Cov}(I_t) = \Sigma_t$ and $\mathrm{Cov}(\dot I_t, I_t) = \dot\alpha_t\alpha_t C_0 + \dot\beta_t\beta_t C_1$. The Gaussian conditioning formula gives, for $x\in\mathrm{Ran}(\Sigma_t)$,
\[
b_t(x) = B(t)\,x,\qquad B(t) = (\dot\alpha_t \alpha_t C_0 + \dot\beta_t \beta_t C_1)\,\Sigma_t^{-1}.
\]
This defines $B(t)$ on the dense domain $\mathrm{Ran}(\Sigma_t)$. We first show it extends to a bounded operator on all of $H$ for every $t\in(0,1)$, then track its norm $\|B(t)\|$ (the Lipschitz constant of the drift, which controls the conditioning of the sampling ODE) as $t\to0$.

\medskip\noindent\textbf{$B(t)$ is diagonal and bounded on $H$ for every $t\in(0,1)$.}~
Both factors $\dot\alpha_t\alpha_t C_0 + \dot\beta_t\beta_t C_1$ and $\Sigma_t^{-1}$ are diagonal in $\{e_j\}$, so $B(t)$ is diagonal as well, $B(t)e_j = b_j(t)\,e_j$, with multipliers
\[
b_j(t) = \frac{\dot\alpha_t\alpha_t\, c_0(j) + \dot\beta_t\beta_t\, c_1(j)}{\alpha_t^2\, c_0(j) + \beta_t^2\, c_1(j)} = g_t(\lambda_j),\qquad \lambda_j := \frac{c_1(j)}{c_0(j)},\quad g_t(\lambda) := \frac{\dot\alpha_t\alpha_t + \dot\beta_t\beta_t\,\lambda}{\alpha_t^2 + \beta_t^2\,\lambda}.
\]
A diagonal operator is bounded on $H$ if and only if its multipliers are bounded, with operator norm equal to their supremum. On $[0,\infty)$ the scalar $g_t$ is continuous and monotone, running between $g_t(0) = \dot\alpha_t/\alpha_t$ and the limit $g_t(\lambda)\to\dot\beta_t/\beta_t$ as $\lambda\to\infty$; in particular $|g_t|\le\max(|\dot\alpha_t/\alpha_t|,\,|\dot\beta_t/\beta_t|)$ there. Since each $\lambda_j\in[0,\infty)$, $B(t)$ extends from $\mathrm{Ran}(\Sigma_t)$ to a bounded operator on $H$, with
\[
\|B(t)\| = \sup_j |b_j(t)| = \sup_j |g_t(\lambda_j)| \;\le\; \max\!\Big(\Big|\tfrac{\dot\alpha_t}{\alpha_t}\Big|,\ \Big|\tfrac{\dot\beta_t}{\beta_t}\Big|\Big) \;<\;\infty,\qquad t\in(0,1),
\]
whether or not the ratios $\lambda_j = c_1(j)/c_0(j)$ are bounded.

\medskip\noindent\textbf{Behavior as $t\to0$.}~
Set $\lambda^\sharp := \sup_j \lambda_j = \sup_j c_1(j)/c_0(j) \in [0,\infty]$. Since $\|B(t)\|$ is the supremum of $|g_t|$ over the multiplier values $\{\lambda_j\}$, the two regimes differ only in how far these values reach.

\emph{If $\lambda^\sharp<\infty$} (equivalently, $C_1 C_0^{-1}$ admits a bounded extension to $H$), then all $\lambda_j$ lie in the bounded set $[0,\lambda^\sharp]$. As $t\to0$ we have $\beta_t\to0$, so $g_t(\lambda)\to\dot\alpha_0/\alpha_0$ uniformly on $[0,\lambda^\sharp]$; hence $B(t)\to(\dot\alpha_0/\alpha_0)\,\mathrm I$ in norm and $\|B(t)\|$ remains bounded.

\emph{If $\lambda^\sharp=\infty$}, then the $\lambda_j$ are unbounded. Since $g_t(\lambda)\to\dot\beta_t/\beta_t$ as $\lambda\to\infty$, this forces $\|B(t)\| = \sup_j|g_t(\lambda_j)| \ge \dot\beta_t/\beta_t$. The schedule satisfies $\beta_0=0$, $\beta\in C^1([0,1])$, and $\dot\beta_t>0$, so $\dot\beta_t/\beta_t\to\infty$ as $t\to 0^+$; hence $\|B(t)\|\to\infty$. The drift field thus has Lipschitz constant blowing up near the initial time.

\end{proof}

\subsection{Proof for Proposition \ref{prop-gaussian-design-schedule}}
\begin{proof}
    Similar to the proof for Proposition \ref{prop-gaussian-measure}, we have the formula
    \[b_t(x) = \bE[\dot I_t | I_t = x] = \mathrm{Cov}(\dot{I}_t,I_t)\mathrm{Cov}(I_t)^{-1}x = (\alpha_t\dot{\alpha}_tC_0  + \beta_t \dot{\beta}_t C_1)(\alpha_t^2C_0  + \beta_t^2 C_1)^{-1}x\, .\]
    We can calculate the $2$-norm using the eigenvalues:
    \[\|\nabla b_t(x)\|_2 = \max_{1\leq j \leq d} \left|\frac{\alpha_t\dot{\alpha}_t  + \beta_t \dot{\beta}_t \lambda_j}{\alpha_t^2  + \beta_t^2 \lambda_j}\right| = \max_{1\leq j \leq d} \frac{\beta_t \dot{\beta}_t (1-\lambda_j)}{1-\beta_t^2 (1-\lambda_j)} \, ,  \]
    where the last equality follows from $\alpha_t^2 + \beta_t^2 = 1$, which gives the numerator $\alpha_t\dot\alpha_t + \beta_t\dot\beta_t\lambda_j = -\beta_t\dot\beta_t(1-\lambda_j)$; after taking the absolute value and using $\beta_t\dot\beta_t \geq 0$, $\lambda_j \leq 1$, and $0 < \beta_t^2(1-\lambda_j) \leq 1$, both the numerator and the denominator are nonnegative, yielding the stated expression.

    The function $\lambda \to \frac{\beta_t \dot{\beta}_t (1-\lambda)}{1-\beta_t^2 (1-\lambda)}$ is non-increasing for $0 < \lambda \leq 1$. Thus \[\|\nabla b_t(x)\|_2 = \frac{\beta_t \dot{\beta}_t (1-\lambda^\star)}{1-\beta_t^2 (1-\lambda^\star)}\, .\]

    Using the formula \eqref{eqn-high-D-gaussian-alpha-beta}, we get $\beta_t^2 =   \frac{(\lambda^\star)^t - 1}{\lambda^\star-1}$ and thus
    \[\|\nabla b_t(x)\|_2 = \frac{1}{2}|\log \lambda^\star|\, .\]
    In fact, the choice \eqref{eqn-high-D-gaussian-alpha-beta} minimizes the averaged squared 2-norm of the gradient over all $\beta_t$; see \cite{chen2025lipschitz}.
\end{proof}

\section{Proof for General Target Measures}
\label{sec-Proof for General Target Measures}

For a general (non-Gaussian) target $\mu^*$ the drift has no closed form, and its regularity is studied
through the Cameron--Martin space of the noise. Throughout, $V := C_0^{1/2}(H)$ is that space, a Hilbert
space with $\langle y, z\rangle_V := \langle C_0^{-1/2}y, C_0^{-1/2}z\rangle_H$, and we assume as in
Proposition~\ref{prop-lip-CM-space} that $\mu^*$ is supported in the $V$-ball $\{\|y\|_V\le R\}$. Since
$\sfN(0,C_0)$ lives on $H\setminus V$ almost surely, the noise is rougher than every point of
$\mathrm{supp}\,\mu^*$.

We proceed in order: a careful meaning for $m_t(x)$ when $x\in H$ (Section~\ref{subsec-mt-interpretation});
a Bayes formula for the conditional law (Section~\ref{subsec-bayes}); and the directional derivative of
$m_t$, which is a conditional covariance (Section~\ref{subsec-cov}). These give the Lipschitz estimate of
Proposition~\ref{prop-lip-CM-space} (Section~\ref{subsec-proof-lip}); well-posedness
(Proposition~\ref{prop-well-posedness}) then follows by removing the linear part of the drift and solving
the rest on $V$ (Section~\ref{sec-Proof-well-posedness}).

\subsection{Interpreting the denoiser \texorpdfstring{$m_t(x)$}{m\_t(x)} for \texorpdfstring{$x\in H$}{x in H}}
\label{subsec-mt-interpretation}

The denoiser is $m_t(x) := \bE[x_1\mid I_t = x]$; we first settle over which $x$ it must be defined. The
interpolant is $I_t = \alpha_t z + \beta_t x_1$ with $z\sim\sfN(0,C_0)$, so for $t<1$ the noise term is
present, $z\in H\setminus V$ almost surely, and $\mu_t = \mathrm{Law}(I_t)$ charges $H\setminus V$: the
conditioning value $x$ is generically a \emph{rough} element of $H$. Since $x_1\in V$ almost surely,
however, the conditional law of $x_1$ given $I_t = x$ is supported on $V$, and its mean $m_t(x)$ is
$V$-valued.

The one delicate ingredient below is the pairing $\langle y, x\rangle_V$ with $y\in V$ but $x\in H$.
Classically $\langle y, x\rangle_V = \langle C_0^{-1/2}y, C_0^{-1/2}x\rangle_H$ requires $x\in V$; for
$x\in H\setminus V$ it extends to a measurable linear functional on $H$, the centered \emph{Wiener
integral} of $y$ \cite{bogachev1998gaussian,hairer2009introduction}.

\medskip\noindent\textbf{The Wiener-integral extension.}~ In the eigenbasis $C_0 e_n = \lambda_n e_n$
($\lambda_n>0$), with $y_n = \langle y, e_n\rangle_H$ and $x_n = \langle x, e_n\rangle_H$, the inner
product for $y,x\in V$ is
\begin{equation}
\label{eqn-V-inner-series}
\langle y, x\rangle_V = \sum_n \frac{y_n\,x_n}{\lambda_n},
\end{equation}
convergent by Cauchy--Schwarz since $\|y\|_V^2 = \sum_n y_n^2/\lambda_n$ and
$\|x\|_V^2 = \sum_n x_n^2/\lambda_n$ are finite. For $x\in H\setminus V$ it need not converge pointwise;
but under $x\sim\sfN(0, sC_0)$ with $s>0$, the summands $y_n x_n/\lambda_n$ are independent centered
Gaussians of variance $s\,y_n^2/\lambda_n$, summing to $s\|y\|_V^2<\infty$, so the series converges a.s.\
and in $L^2$ to a centered Gaussian
\[
\cI_y : H \to \bR, \qquad \cI_y(x)\sim\sfN\!\big(0,\, s\|y\|_V^2\big).
\]
Thus $\cI_y$ is a measurable linear functional, defined $\sfN(0,sC_0)$-a.e., agreeing with
$\langle y,\cdot\rangle_V$ on $V$, with $\|\cI_y\|_{L^2(\sfN(0,sC_0))} = \sqrt{s}\,\|y\|_V$. Fixing a
version of $(y,x)\mapsto\cI_y(x)$ jointly measurable on $V\times H$ \cite{bogachev1998gaussian}, we write
$\langle y, x\rangle_V := \cI_y(x)$ for $y\in V$, $x\in H$; it reduces to the classical inner product when
$x\in V$.

\subsection{The Bayes formula for the conditional law}
\label{subsec-bayes}

We identify $\mathrm{Law}(x_1\mid I_t = x)$, $\mu_t$-a.e., and read off the denoiser as its mean.

In finite dimensions this is ``posterior $\propto$ likelihood $\times$ prior,'' all densities against
Lebesgue measure. On infinite dimensional $H$ there is no Lebesgue measure and $\mu_t$ has no density, so we run the
computation relative to one fixed Gaussian reference.

\medskip\noindent\textbf{Step 1: likelihood relative to a reference.}~ Conditioning on $x_1 = y$,
$\gamma_y := \mathrm{Law}(I_t\mid x_1 = y) = \sfN(\beta_t y, \alpha_t^2 C_0)$; write
$\gamma_0 := \sfN(0,\alpha_t^2 C_0)$ for the $y$-independent reference (the law of $\alpha_t z$). By the
Cameron--Martin theorem \cite{bogachev1998gaussian}, for a centered Gaussian $\sfN(0,Q)$ and a shift
$h\in Q^{1/2}(H)$,
\[
\frac{\rd\,\sfN(h,Q)}{\rd\,\sfN(0,Q)}(x) = \exp\!\Big(\langle h, x\rangle_Q - \tfrac12\|h\|_Q^2\Big),
\qquad \|h\|_Q^2 := \|Q^{-1/2}h\|_H^2,
\]
with $\langle h,\cdot\rangle_Q$ the Wiener-integral extension of $x\mapsto\langle Q^{-1/2}h, Q^{-1/2}x\rangle_H$.
The shift $h=\beta_t y$ lies in $V$, the Cameron--Martin space of $\gamma_0$; with $Q=\alpha_t^2 C_0$ one
gets $\|h\|_Q^2=\beta_t^2\|y\|_V^2/\alpha_t^2$ and $\langle h,x\rangle_Q=\beta_t\langle y,x\rangle_V/\alpha_t^2$,
so
\begin{equation}
\label{eqn-CM-likelihood}
\frac{\rd\gamma_y}{\rd\gamma_0}(x) = e^{\Phi_t(x,y)},
\qquad
\Phi_t(x,y) := \frac{\beta_t}{\alpha_t^2}\langle y, x\rangle_V - \frac{\beta_t^2}{2\alpha_t^2}\|y\|_V^2 .
\end{equation}
The Wiener integral in the linear term is what makes $\Phi_t(x,y)$ well defined $\gamma_0$-a.e.

\medskip\noindent\textbf{Step 2: joint law against a product reference.}~ Since $x_1\sim\mu^*$ and
$I_t\mid x_1=y\sim\gamma_y$, for bounded measurable $f$ on $V\times H$,
\[
\int f\,\rd\,\mathrm{Law}(x_1,I_t)
= \int_V\!\!\int_H f(y,x)\,\gamma_y(\rd x)\,\mu^*(\rd y)
= \int_V\!\!\int_H f(y,x)\,e^{\Phi_t(x,y)}\,\gamma_0(\rd x)\,\mu^*(\rd y)
\]
by~\eqref{eqn-CM-likelihood}. Hence the joint law has density
\begin{equation}
\label{eqn-joint-density}
\frac{\rd\,\mathrm{Law}(x_1,I_t)}{\rd(\mu^*\otimes\gamma_0)}(y,x) = e^{\Phi_t(x,y)}
\end{equation}
against the product reference $\mu^*\otimes\gamma_0$.

\medskip\noindent\textbf{Step 3: marginal and posterior.}~ Integrating~\eqref{eqn-joint-density} over $y$
gives the marginal of $I_t$ as a density against the same reference,
\begin{equation}
\label{eqn-marginal-density}
\mu_t(\rd x) = Z_t(x)\,\gamma_0(\rd x),
\qquad
Z_t(x) := \int_V e^{\Phi_t(x,y)}\,\mu^*(\rd y).
\end{equation}
Disintegrating~\eqref{eqn-joint-density} against this marginal gives
\begin{equation}
\label{eqn-cond-law}
\mathrm{Law}(x_1\mid I_t=x)(\rd y) = \frac{e^{\Phi_t(x,y)}}{Z_t(x)}\,\mu^*(\rd y),
\end{equation}
the same conditional law as in the bounded-Cameron--Martin-support setting of
\cite[Theorem~12]{pidstrigach2023infinite}. Because $\mu^*$ has bounded $V$-support, $\|y\|_V\le R$ holds $\mu^*$-a.s.; on $\{\|y\|_V\le R\}$ the quadratic term of $\Phi_t$ is then bounded and the linear term is integrable against the Gaussian reference $\gamma_0$, so Tonelli's theorem gives
$Z_t(x)\in(0,\infty)$ for $\gamma_0$-a.e.\ (hence $\mu_t$-a.e.) $x$ and~\eqref{eqn-cond-law} is a probability measure on
$\{\|y\|_V\le R\}$. The denoiser is its mean,
\begin{equation}
\label{eqn-denoiser}
m_t(x) = \frac{1}{Z_t(x)}\int_V y\,e^{\Phi_t(x,y)}\,\mu^*(\rd y),
\end{equation}
a barycenter over the convex set $\{\|y\|_V\le R\}$, so $\|m_t(x)\|_V\le R$ --- which proves part~(i) of
Proposition~\ref{prop-lip-CM-space}.

\subsection{The directional derivative of \texorpdfstring{$m_t$}{m\_t} as a conditional covariance}
\label{subsec-cov}

Fix $x, w\in V$ and a direction $u\in V$, and differentiate $s\mapsto\langle m_t(x + sw), u\rangle_V$.
From~\eqref{eqn-denoiser}, $\langle m_t(x+sw), u\rangle_V = N(s)/Z(s)$ with
\[
N(s) := \int_V \langle y, u\rangle_V\, e^{g(s;y)}\,\mu^*(\rd y),
\qquad
Z(s) := \int_V e^{g(s;y)}\,\mu^*(\rd y),
\]
\[
g(s;y) := \Phi_t(x+sw,y) = \Phi_t(x,y) + s\,\kappa_t\langle y, w\rangle_V,
\qquad \kappa_t := \frac{\beta_t}{\alpha_t^2}.
\]
Since $x,w\in V$, the pairings are classical inner products ($\mu^*$-a.e.). As $\partial_s g(s;y) = \kappa_t\langle y,w\rangle_V$ is bounded by $\kappa_t R\|w\|_V$ on
$\{\|y\|_V\le R\}$, differentiation under the integral is justified (dominated convergence) and $Z(s)>0$:
\[
N'(s) = \kappa_t\!\int_V \langle y, u\rangle_V\langle y, w\rangle_V\, e^{g(s;y)}\,\mu^*(\rd y),
\qquad
Z'(s) = \kappa_t\!\int_V \langle y, w\rangle_V\, e^{g(s;y)}\,\mu^*(\rd y).
\]
Let $\nu_s(\rd y) := e^{g(s;y)}\mu^*(\rd y)/Z(s)$, the conditional law~\eqref{eqn-cond-law} at $x+sw$,
carried by $\{\|y\|_V\le R\}$. Then $N/Z = \bE_{\nu_s}[\langle y,u\rangle_V]$,
$N'/Z = \kappa_t\bE_{\nu_s}[\langle y,u\rangle_V\langle y,w\rangle_V]$, and
$Z'/Z = \kappa_t\bE_{\nu_s}[\langle y,w\rangle_V]$, so the quotient rule gives
\[
\frac{\rd}{\rd s}\frac{N(s)}{Z(s)}
= \frac{N'(s)}{Z(s)} - \frac{N(s)}{Z(s)}\cdot\frac{Z'(s)}{Z(s)}
= \kappa_t\Big(\bE_{\nu_s}\big[\langle y,u\rangle_V\langle y,w\rangle_V\big]
- \bE_{\nu_s}\big[\langle y,u\rangle_V\big]\bE_{\nu_s}\big[\langle y,w\rangle_V\big]\Big),
\]
that is,
\begin{equation}
\label{eqn-deriv-cov}
\frac{\rd}{\rd s}\langle m_t(x + sw), u\rangle_V
= \kappa_t\,\mathrm{Cov}_{\nu_s}\!\big(\langle y, u\rangle_V,\, \langle y, w\rangle_V\big).
\end{equation}
The directional derivative of the denoiser is thus the conditional covariance of the linear statistics
$\langle y, u\rangle_V$ and $\langle y, w\rangle_V$, scaled by $\kappa_t$. Since $\nu_s$ is carried by
$\{\|y\|_V\le R\}$, every such statistic obeys $|\langle y, u\rangle_V|\le R\|u\|_V$, so Cauchy--Schwarz
gives
\begin{equation}
\label{eqn-cov-bound}
\big|\mathrm{Cov}_{\nu_s}(\langle y, u\rangle_V, \langle y, w\rangle_V)\big|
\le \sqrt{\bE_{\nu_s}\langle y,u\rangle_V^2}\,\sqrt{\bE_{\nu_s}\langle y,w\rangle_V^2}
\le R^2\,\|u\|_V\,\|w\|_V .
\end{equation}

\subsection{Proof of Proposition~\ref{prop-lip-CM-space}}
\label{subsec-proof-lip}

\begin{proof}
Part~(i), the bound $\|m_t(x)\|_V\le R$, was established after~\eqref{eqn-denoiser}. The drift
representation~\eqref{eqn-bt-denoiser-form} is immediate from linearity: since
$\dot I_t = \dot\alpha_t z + \dot\beta_t x_1$ and $\alpha_t z = I_t - \beta_t x_1$,
\begin{equation}
\label{eqn-bt-rep}
b_t(x) = \dot\alpha_t\,\bE[z\mid I_t = x] + \dot\beta_t\, m_t(x)
= \tfrac{\dot\alpha_t}{\alpha_t}\big(x - \beta_t m_t(x)\big) + \dot\beta_t\, m_t(x)
= \tfrac{\dot\alpha_t}{\alpha_t}\,x + c(t)\, m_t(x),
\end{equation}
with $c(t) := \dot\beta_t - \tfrac{\beta_t\dot\alpha_t}{\alpha_t}$.

For part~(ii), fix $x, w\in V$. For any $u\in V$, the fundamental theorem of calculus
and~\eqref{eqn-deriv-cov} give
\[
\langle m_t(x + w) - m_t(x),\, u\rangle_V
= \kappa_t\!\int_0^1 \mathrm{Cov}_{\nu_s}\!\big(\langle y, u\rangle_V, \langle y, w\rangle_V\big)\,\rd s,
\]
and~\eqref{eqn-cov-bound} bounds the right-hand side by $\kappa_t R^2\|w\|_V\|u\|_V$. Taking the supremum
over $\|u\|_V\le 1$ shows that $m_t\colon V\to V$ is Lipschitz,
\begin{equation}
\label{eqn-mt-lip}
\|m_t(x + w) - m_t(x)\|_V \le \tilde L_t\,\|w\|_V,
\qquad
\tilde L_t := \kappa_t R^2 = \frac{\beta_t R^2}{\alpha_t^2}.
\end{equation}
By~\eqref{eqn-bt-rep}, $b_t(x+w) - b_t(x) = \tfrac{\dot\alpha_t}{\alpha_t}\,w + c(t)\big(m_t(x+w) - m_t(x)\big)$,
so $b_t\colon V\to V$ is Lipschitz with constant
\[
L_t := \Big|\tfrac{\dot\alpha_t}{\alpha_t}\Big| + |c(t)|\,\tilde L_t .
\]
Both $\tilde L_t$ and $L_t$ are finite for $t<1$, with $\sup_{t\in[0,1-\delta]}L_t<\infty$ (as
$\alpha_t\ge\alpha_{1-\delta}>0$ and the schedule is $C^1$); the bound depends only on $\delta$, $R$, and
the schedule.
\end{proof}

\subsection{Proof of Proposition~\ref{prop-well-posedness}}
\label{sec-Proof-well-posedness}

\begin{proof}
$\delta\in(0,1)$ is fixed.

\medskip\noindent\textbf{Case (a): Gaussian.}~ Here $b_t(x) = B(t)x$ with
$B(t) = (\dot\alpha_t\alpha_t C_0 + \dot\beta_t\beta_t C_1)\Sigma_t^{-1}$
(Proposition~\ref{prop-gaussian-measure}). When $C_1 C_0^{-1}$ is bounded, that proof gives
$\sup_{t\in[0,1-\delta]}\|B(t)\|_{H\to H}<\infty$, so $b_t$ is globally Lipschitz on $H$ uniformly in $t$;
Picard--Lindel\"of on $H$ yields a unique $X_t = U(t,0)X_0\in C^1([0,1-\delta];H)$ for every $X_0$, where $U$ is
the evolution family of $B(\cdot)$.

\medskip\noindent\textbf{Case (b): bounded $V$-support.}~ Since $m_t$ is Lipschitz only along $V$, we
remove the linear part of the drift by variation of constants and solve the rest on $V$.

\smallskip\noindent\emph{Step 1 (variation of constants).}~ Fix $X_0 = a\in H$. Multiplying
$\dot X_t = b_t(X_t)$ by the integrating factor $1/\alpha_t$ of the linear part
$\tfrac{\dot\alpha_t}{\alpha_t}x$ and integrating (with $\alpha_0=1$), any $H$-valued solution satisfies
\begin{equation}
\label{eqn-VOC}
X_t = \alpha_t a + \alpha_t\int_0^t \frac{c(s)}{\alpha_s}\, m_s(X_s)\,\rd s.
\end{equation}
As $m_s(X_s)\in V$ with $\|m_s(X_s)\|_V\le R$ and $c(s)/\alpha_s$ is bounded on $[0,1-\delta]$, the integral
is a $V$-valued Bochner integral. Hence every solution from $a$ has $X_t - \alpha_t a\in V$: it stays on
the slice $\alpha_t a + V$, the $H$-component being the explicit scaling $\alpha_t a$.

\smallskip\noindent\emph{Step 2 (correction on $V$).}~ The correction $W_t := X_t - \alpha_t a\in V$ solves
\begin{equation}
\label{eqn-Wt-ode}
\dot W_t = \tfrac{\dot\alpha_t}{\alpha_t}\, W_t + c(t)\, m_t\big(W_t + \alpha_t a\big), \qquad W_0 = 0,
\end{equation}
an ODE on $V$ with nonlinearity the denoiser along the slice.

\smallskip\noindent\emph{Step 3 (the correction equation is Lipschitz on $V$).}~ We show the nonlinear part
$w\mapsto m_t(w+\alpha_t a)$ of~\eqref{eqn-Wt-ode} is Lipschitz $V\to V$ with the constant $\tilde L_t$
of~\eqref{eqn-mt-lip}, for $\sfN(0,C_0)$-a.e.\ $a$. At $x = w+\alpha_t a$, linearity of the Wiener integral
gives $\langle y, w+\alpha_t a\rangle_V = \langle y,w\rangle_V + \alpha_t\langle y,a\rangle_V$, so the
exponent $\Phi_t(w+\alpha_t a,y)$ in~\eqref{eqn-cond-law} is $\kappa_t\langle y,w\rangle_V$ plus a
$w$-independent remainder, which reweights $\mu^*$ into
\[
\mu^*_a(\rd y)\ \propto\ \exp\!\Big(-\tfrac{\beta_t^2}{2\alpha_t^2}\|y\|_V^2
+ \tfrac{\beta_t}{\alpha_t}\langle y,a\rangle_V\Big)\mu^*(\rd y),
\]
still carried by $\{\|y\|_V\le R\}$. Its mass is finite for $\sfN(0,C_0)$-a.e.\ $a$, simultaneously for all
$t\in[0,1-\delta]$: with $c := \sup_{t\in[0,1-\delta]}\beta_t/\alpha_t<\infty$ it is bounded by
$1+\int_V e^{c\langle y,a\rangle_V}\mu^*(\rd y)$, which is finite a.e.\ by Tonelli and the Gaussian Laplace
transform $\bE_{a\sim\sfN(0,C_0)}e^{c\langle y,a\rangle_V} = e^{c^2\|y\|_V^2/2}$:
\[
\int_H\!\int_V e^{c\langle y,a\rangle_V}\,\mu^*(\rd y)\,\sfN(0,C_0)(\rd a)
= \int_V e^{c^2\|y\|_V^2/2}\,\mu^*(\rd y) \le e^{c^2 R^2/2}<\infty.
\]
For such $a$, $m_t(\,\cdot+\alpha_t a)$ is the mean of the $\kappa_t\langle\cdot,w\rangle_V$-tilt of
$\mu^*_a$, so the computation of Section~\ref{subsec-cov} applies with $\mu^*$ replaced by $\mu^*_a$; as
\eqref{eqn-cov-bound} uses only the radius $R$, the Lipschitz constant is $\tilde L_t$, independent of $a$.

Hence the right-hand side of~\eqref{eqn-Wt-ode} is Lipschitz on $V$ with constant
$L_t = |\dot\alpha_t/\alpha_t| + |c(t)|\,\tilde L_t$ (Section~\ref{subsec-proof-lip}), uniformly bounded on
$[0,1-\delta]$. Picard--Lindel\"of on $V$ gives a unique $W_\cdot\in C^1([0,1-\delta];V)$ with $W_0=0$, and
$X_t := W_t + \alpha_t a$ is the unique $H$-valued solution from $a$ (Step~1). It is measurable in $a$ (each
Picard iterate is, through the jointly measurable $\cI_y$ in $\mu^*_a$), so $X_t = T_t(X_0)$ is a measurable
flow, defined $\sfN(0,C_0)$-a.e.

\smallskip\noindent\emph{Step 4 (matching the marginals).}~ It remains to verify the second hypothesis of
Proposition~\ref{prop-si-bt-functionspace}: that the interpolant marginals are carried by the flow,
$\mathrm{Law}(X_t)=\mu_t$. The proof of that proposition already shows $\mu_t$ to be a weak solution of the
continuity equation $\partial_t\mu_t+\nabla\!\cdot(b_t\mu_t)=0$; the hypothesis asks, in addition, that this
weak solution be unique. We obtain the identification---and with it the uniqueness---from the superposition
principle \cite{ambrosio2014continuity}, in the metric-measure-space form valid on the separable Hilbert
space $H$: a continuously varying family of probability measures that solves the continuity equation for a
drift integrable along the way is realized as the distribution of trajectories of
$\dot\gamma_t=b_t(\gamma_t)$. The drift is integrable along the marginals,
\[
\int_0^{1-\delta}\!\int_H\|b_t(x)\|_H\,\mu_t(\rd x)\,\rd t<\infty ,
\]
so $\mu_t$ is carried by solutions of $\dot\gamma_t=b_t(\gamma_t)$. Since this ODE has a unique solution from
$\sfN(0,C_0)$-a.e.\ starting point---by Picard--Lindel\"of on $H$ in Case~(a), and by Steps~1--3 in
Case~(b)---those trajectories are the single flow $X_t=T_t(X_0)$. Hence
$\mu_t=(T_t)_\#\sfN(0,C_0)=\mathrm{Law}(X_t)$ for $t\in[0,1-\delta]$, which is therefore the unique weak
solution.
\end{proof}

\section{Experimental Details}
\label{appendix-reproducibility}

This appendix collects the data-generation, training, and evaluation details that are needed to reproduce all numerical results reported in the main text. 

\medskip\noindent\textbf{Synthetic Gaussian field.}
The target measure is $\sfN(0, C_1)$ with $C_1 = \sigma_1^2(-\Delta+\tau_1^2 \mathrm{I})^{-s_1}$ on $D=[0,1]^2$ with $s_1=3$, $\tau_1=1$, $\sigma_1^2=(4\pi^2+\tau_1^2)^{s_1}$. Test samples are drawn directly via the explicit formula in Section~\ref{sec-Example: Matern fields} ($5\times 10^4$ i.i.d.\ samples per resolution for spectrum estimation). The white noise is i.i.d.\ standard normal per pixel; the matched-spectrum noise uses the same explicit formula with $C_0=C_1$. The drift $b_t$ is evaluated directly in closed form via Proposition~\ref{prop-gaussian-measure} (mode-wise diagonal action implemented by FFT); no neural network is trained. The same closed-form drift is used in all Gaussian-case experiments (Section~\ref{sec-example-Gaussian} and Appendix~\ref{appendix-validate-prop33}.B--C).

\medskip\noindent\textbf{Allen--Cahn invariant distribution.}
We integrate the stochastic Allen--Cahn equation on $[0,1]$ with finite differences on $N$ equispaced grid points and double-well potential $V(u)=(1-u^2)^2$. Samples from the invariant measure are obtained by the ensemble preconditioned MCMC of \cite{chen2025new}, giving $5\times 10^4$ training samples after burn-in. The matched-spectrum noise matches the Gaussian component $\exp(-\int_0^1 \tfrac12(\partial_x u)^2\mathrm{d}x)$, sampled via fast Fourier transform. The drift is learned under the linear schedule with AdamW (batch size $100$, learning rate $2\times 10^{-4}$ with cosine decay, $5\times 10^4$ gradient steps, gradient-norm clipping at $10^4$). The UNet (with 1D convolutions) has base channels $32$, multipliers $(1,2,2,2)$, four downsampling/upsampling stages, four attention heads at the lower-resolution stages, and a learned sinusoidal time embedding of dimension $32$, giving $\approx 2$M parameters.

\medskip\noindent\textbf{Navier--Stokes invariant distribution.}
We integrate the stochastically forced Navier--Stokes vorticity equation \eqref{eq:2D_vorticity_NS} on the torus $\mathbb T^2 = [0,2\pi]^2$ with a pseudo-spectral solver, explicit Heun time-stepping, $2/3$ de-aliasing, $\nu = 10^{-3}$, $\alpha = 0.1$, $\varepsilon = 1$, and the low-frequency forcing of \cite{chen2024probabilistic}. We run five independent trajectories on a $256\times 256$ grid and, after a long burn-in, save vorticity snapshots at fixed intervals, yielding approximately $10^5$ snapshots in total ($90\%$ training / $10\%$ test). Each snapshot is normalized by a fixed empirical per-pixel norm to unit standard deviation; for evaluation at $128\times 128$, $64\times 64$, or $32\times 32$, snapshots are bilinearly downsampled from $256\times 256$.

The white noise is i.i.d.\ standard normal per pixel; the matched-spectrum noise uses the empirical per-mode variance estimated from $5\times 10^3$ training snapshots; the rougher (mul-$k$) noise multiplies the matched per-mode standard deviation by $k=\|m\|_2$. Training uses the same $\approx 2$M-parameter UNet and AdamW protocol as the Allen--Cahn setting (with 2D convolutions).

\medskip\noindent\textbf{Sampling and seeds.}
At inference time the ODE is integrated by fixed-step RK4 on a uniform grid in $[t_{\min}, t_{\max}] = [10^{-3}, 1-10^{-3}]$. For the designed schedule, the drift is obtained at inference time from the trained linear-schedule drift via the transfer formula~\cite{chen2025lipschitz, kingma2021variational, karras2022elucidating}, requiring no retraining. For each (schedule, step count, seed) configuration we generate $500$ samples from a fixed batch of initial conditions, shared across schedules within a seed for paired comparison. Means and standard deviations are reported across five independent random seeds.

\medskip\noindent\textbf{Convergence to a fixed loss tolerance.}
All models were trained to within $1\%$ of the loss minimum observed over $5\times 10^4$ gradient steps; longer training (up to $1.5\times 10^5$ steps) did not change the spectrum-error metrics within standard deviation.

\medskip\noindent\textbf{Spectra and band errors.}
For a 2D field $\omega$ of side $N$, $\hat\omega(\mathbf m) := \mathcal F\omega$, and the radially-averaged enstrophy spectrum is
\begin{equation*}
    S(k) = \pi(k_+^2-k_-^2)\cdot \mathrm{mean}\{|\hat\omega(\mathbf m)|^2 : |\mathbf m|\in[k_-,k_+)\},
\end{equation*}
where $(k_-,k_+)=(k-0.5,k+0.5)$ and $k$ ranges over integer wavenumbers from $1$ to $N/2$. Per-band relative errors are unweighted averages of $|S_{\rm gen}(k)-S_{\rm truth}(k)|/|S_{\rm truth}(k)|$ over $k$ in each band; band cuts are $k<8$, $8\le k<24$, $k\ge 24$.

\medskip\noindent\textbf{Non-Gaussian statistics.}
For a snapshot $\omega$, structure functions are
\begin{equation*}
    S_p(r) = \tfrac12\,\mathbb E[|\omega(\cdot+re_1)-\omega(\cdot)|^p] + \tfrac12\,\mathbb E[|\omega(\cdot+re_2)-\omega(\cdot)|^p],
\end{equation*}
flatness $F(r)=S_4(r)/S_2(r)^2$, and the gradient kurtosis equals $F(1)$.

\section{Numerical Validation of Propositions}
\label{appendix-validate-prop33}
This appendix provides numerical evidence for the analytical claims of the three propositions: the rougher-than-data condition (\ref{appendix-validate-prop33}.A); the closed-form drift Lipschitz constants of Propositions~\ref{prop-gaussian-measure} and~\ref{prop-gaussian-design-schedule} (\ref{appendix-validate-prop33}.B); and the wavenumber-dependent versus scalar variant of the designed schedule (\ref{appendix-validate-prop33}.C).

\subsection*{A.\ Cameron-Martin norm of the data across resolutions}
 The table below reports the empirical mean of $\|x_1\|_V^2$ (over $2000$ samples; relative standard deviations below $1\%$) at $N\in\{32,64,128\}$ for all three settings (synthetic Gaussian, Allen--Cahn, and Navier--Stokes), against three candidate noise distributions per setting.

\begin{center}
\small
\begin{tabular}{lccc}
\toprule
\multicolumn{4}{c}{\bf Synthetic Gaussian (Mat\'ern, $s_1=3$)} \\
Noise $C_0$ & $\|x_1\|_V^2$ at $N{=}32$ & $N{=}64$ & $N{=}128$\\
\midrule
White noise ($s_0=0$, rougher)        & $4.56$ & $4.63$ & $4.55$ \\
Mat\'ern ($s_0=2$, rougher)           & $20.9$ & $25.5$ & $29.9$ \\
Mat\'ern ($s_0=3$, matched)           & $1.02\!\times\!10^{3}$ & $4.09\!\times\!10^{3}$ & $1.64\!\times\!10^{4}$ \\
Mat\'ern ($s_0=4$, smoother than data) & $1.71\!\times\!10^{5}$ & $2.73\!\times\!10^{6}$ & $4.37\!\times\!10^{7}$ \\
\midrule
\multicolumn{4}{c}{\bf Allen--Cahn (1D, $N$ grid points)} \\
White noise (rougher)                 & $0.86$ & $0.85$ & $0.84$ \\
$(-\partial_x^2)^{-1}$ (matched)  & $2.93$ & $4.88$ & $9.00$ \\
$(-\partial_x^2)^{-2}$ (smoother)      & $236$  & $1.77\!\times\!10^{3}$ & $1.41\!\times\!10^{4}$ \\
\midrule
\multicolumn{4}{c}{\bf Navier--Stokes (2D, downsampled from native $256\times 256$)} \\
White noise (much rougher)              & $0.99$ & $1.00$ & $1.00$ \\
Mul-$k$ rougher spectrum noise          & $20.7$ & $25.1$ & $29.4$ \\
Matched-spectrum noise                  & $1.02\!\times\!10^{3}$ & $4.10\!\times\!10^{3}$ & $1.64\!\times\!10^{4}$ \\
\bottomrule
\end{tabular}
\end{center}

The scaling with $N$ separates the noise choices into three regimes. (i) \emph{Strictly rougher than data} ($s_1-s_0>d/2$, i.e.\ $>1$ in 2D and $>1/2$ in 1D; all white-noise rows): $\|x_1\|_V^2$ is stable in $N$, indicating $x_1\in V$ in the continuum limit---a prerequisite for, though weaker than, the bounded-$V$-support hypothesis $\|x_1\|_V\le R$ a.s.\ of Proposition~\ref{prop-lip-CM-space}. The 2D borderline cases at $s_1-s_0=1$ (Mat\'ern $s_0=2$ and the Navier--Stokes mul-$k$ noise) instead grow logarithmically, $\|x_1\|_V^2\sim 2\pi\log(N/2)$ (increments $\approx 2\pi\log 2\approx 4.35$ per doubling, the critical 2D rate), so in the continuum they are marginally \emph{not} in $V$. (ii) \emph{Matched smoothness}: $\|x_1\|_V^2$ grows polynomially with $N$ (linearly in $N$ for 1D, $N^2$ for the 2D). (iii) \emph{Smoother than data}: $\|x_1\|_V^2$ grows even faster (e.g.\ $\sim N^4$ in the Gaussian setting).

\subsection*{B.\ Drift Lipschitz constant for the Gaussian closed form}
\label{sec-app-lip-validation}
\begin{figure}[ht]
    \centering
    \includegraphics[width=0.85\linewidth]{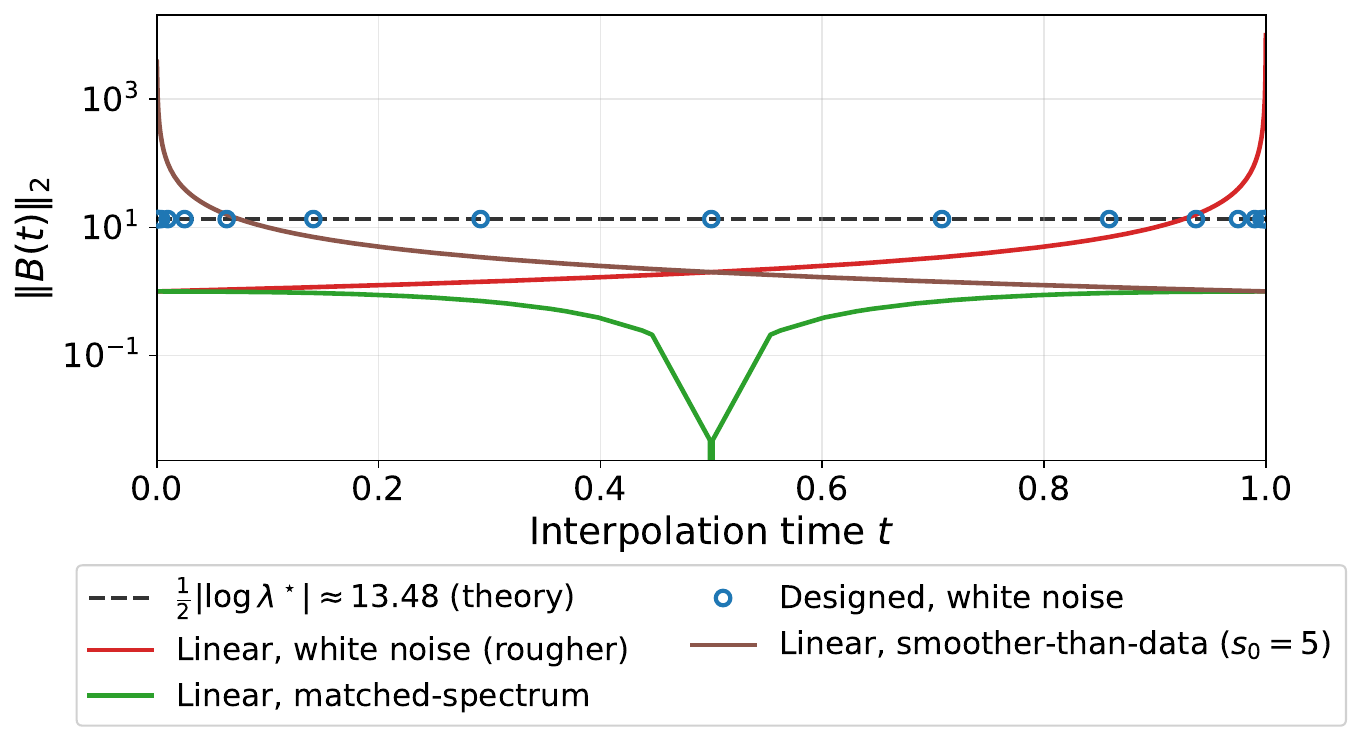}
    \caption{Drift Lipschitz constant $\|B(t)\|_2$ for the closed-form Gaussian drift on $128\times 128$ Mat\'ern data ($s_1=3$). Dashed black line: theoretical designed-schedule bound $\frac{1}{2}|\log\lambda^\star|\approx 13.48$, hit exactly by the designed + white noise curve (open circles). Smoother-than-data noise ($s_0=5$) tracks the schedule's $1/t$ ceiling at small $t$. Matched-spectrum noise is bounded at $\sim 1$. Linear + white noise is stiff near the terminal time.}
    \label{fig:lip-validation}
\end{figure}

For Gaussian targets, the drift $b_t(x)=B(t)x$ has Lipschitz constant equal to $\|B(t)\|_2 = \max_m|\tilde B(t;m)|$ in the Fourier basis (zero mode excluded). Figure~\ref{fig:lip-validation} plots this for four configurations on the $128\times 128$ Mat\'ern-like data ($s_1=3$):
\begin{itemize}
    \item Linear + matched-spectrum: bounded uniformly at $\sim 1$.
    \item Linear + white noise (rougher): bounded near $t=0$, grows as $1/(1-t)$ near $t=1$ (the schedule's terminal-time stiffness, $\sim 100$ at $t=0.99$).
    \item Linear + smoother-than-data ($s_0=5$): tracks the schedule's $1/t$ ceiling at small $t$ ($\sim 100$ at $t=10^{-2}$, $\sim 985$ at $t=10^{-3}$), confirming Proposition~\ref{prop-gaussian-measure}'s prediction of $\|B(t)\|_2\to\infty$ as $t\to 0$.
    \item Designed schedule + white noise: $\|B(t)\|_2$ is constant in $t$ at $\approx 13.48$, agreeing to four significant figures with the analytical bound $\frac{1}{2}|\log\lambda^\star| = 13.479$ predicted by Proposition~\ref{prop-gaussian-design-schedule}.
\end{itemize}

\subsection*{C.\ Wavenumber-dependent vs.\ scalar designed schedule}
\label{sec-app-wavenum-vs-scalar}
\begin{figure}[ht]
    \centering
    \includegraphics[width=0.78\linewidth]{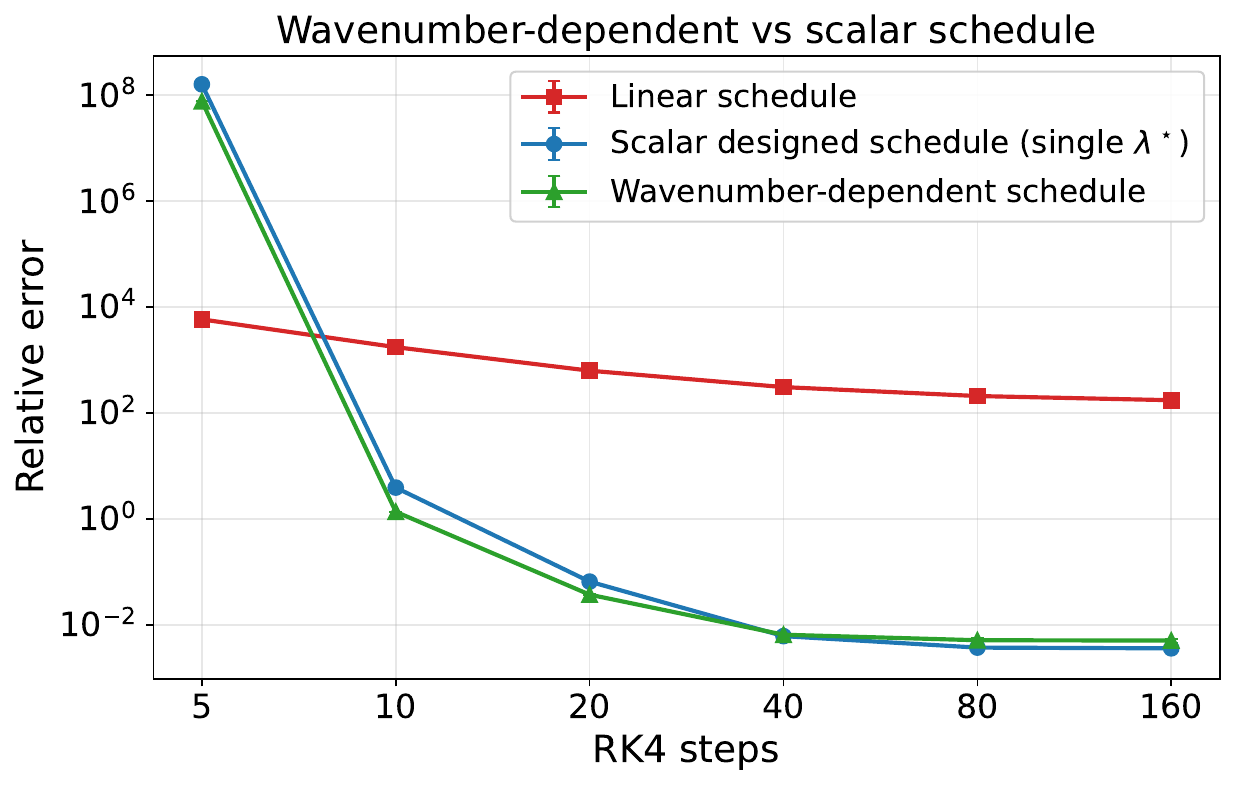}
    \caption{Wavenumber-dependent vs.\ scalar designed schedule on the $128\times 128$ Mat\'ern Gaussian target ($s_1=3$, white noise, $\lambda^\star \approx 1.96\times 10^{-12}$). Horizontal axis: RK4 step count (log scale). Vertical axis: relative high-wavenumber spectrum error (log scale, $k\ge 24$). Closed-form drift, fixed-step RK4 with $t\in[10^{-4}, 1-10^{-4}]$, five independent seeds, $500$ samples per seed; error bars (mostly invisible at this scale) are below $1\%$ relative for entries with $\ge 10$ steps. At $5$ RK4 steps both scaled schedules are inaccurate; at $\ge 10$ steps they outperform the linear schedule by $4$--$6$ orders of magnitude, with the wavenumber-dependent variant ahead by a small constant factor at $10$--$20$ steps; at $\ge 40$ steps both saturate at the accuracy floor and the small spread between them is within seed variation.}
    \label{fig:wavenum-vs-scalar}
\end{figure}

The wavenumber-dependent schedule \eqref{eqn-high-D-gaussian-alpha-beta} with $\lambda^\star$ replaced by $\lambda_m$ per mode is, in principle, the most natural choice in the Gaussian Mat\'ern-like setting: each Fourier mode is then integrated with the schedule that exactly minimizes its drift Lipschitz constant (in the closed-form Gaussian setting this drift is constant in time and equals $\frac12\log(c_1(m)/c_0(m))$ per mode). The scalar schedule used in the main text is a worst-case alternative that uses the smallest $\lambda^\star$ across resolved modes for \emph{all} modes; it is much easier to implement because it does not need to act differently on different Fourier modes during ODE integration. Figure~\ref{fig:wavenum-vs-scalar} compares the two on the $128\times 128$ Mat\'ern target ($s_1=3$, white noise, $\lambda^\star\approx 1.96\times 10^{-12}$ at the highest resolved wavenumber).

Three regimes are visible. (i)~For very few integrator steps ($\le 5$), \emph{both} scaled schedules suffer from poor accuracy. (ii)~In the intermediate-step regime ($10$--$20$), the wavenumber-dependent variant is ahead of the scalar one by a small constant factor (about $2$--$3\times$), reflecting its per-mode optimality. (iii)~At $\ge 40$ steps, both schedules saturate at the accuracy floor of $\sim 3\!\times\!10^{-3}$ and the small spread between the two curves is within the seed-to-seed variation; at this floor the finite-sample error dominates, not the schedule choice. The linear schedule remains a factor $\sim 5\times 10^4$ above this floor at every step count, in agreement with Table~\ref{tab:smoothness-sweep}. The price of the scalar schedule's simplicity is therefore at most a small constant factor in step count in the regime where it matters (low NFE), while the gain over the linear schedule is (at fixed accuracy) exponential in $|\log\lambda^\star|$, for such Gaussian example.

\bibliography{ref}
\bibliographystyle{plain}

\end{document}